\documentclass[referee,iicol,pdflatex,sn-mathphys]{sn-jnl}

\usepackage{times}
\usepackage{soul}
\usepackage{url}
\usepackage{graphicx}
\usepackage{amsmath}
\usepackage{amssymb}
\usepackage{amsthm}
\usepackage{booktabs}
\usepackage[normalem]{ulem}

\usepackage{balance} 
\usepackage{balance} 
\usepackage{times}  
\usepackage{helvet} 
\usepackage{courier}  
\usepackage{comment}
\usepackage{bbm}
\usepackage{xcolor}
\usepackage{multirow}
\usepackage{enumitem}
\usepackage{amsmath}
\usepackage{algpseudocode}
\usepackage{algorithm}
\usepackage{ dsfont } 
\usepackage{subcaption}
\usepackage{siunitx}

\newcount\Comments  
\long\def\commentp#1{{\ifnum\Comments=1\textcolor{red}{\bf --P: #1--}\fi}}
\long\def\commentps#1{{\ifnum\Comments=4\textcolor{red}{\bf --P: #1--}\fi}}
\long\def\ijcaicommentp#1{{\ifnum\Comments=3\textcolor{red}{\bf --P: #1--}\fi}}
\long\def\commentd#1{{\ifnum\Comments=1\textcolor{orange}{\bf --D: #1--}\fi}}
\long\def\commentdu#1{{\ifnum\Comments=4\textcolor{orange}{\bf --D: #1--}\fi}} 
\long\def\changedyz#1{{\ifnum\Comments=5\textcolor{blue}{#1}\else #1\fi}}  
\long\def\changedyzz#1{{\ifnum\Comments=4\textcolor{blue}{#1}\else #1\fi}}  
\long\def\ijcaicommentd#1{{\ifnum\Comments=3\textcolor{orange}{\bf --D: #1--}\fi}} 
\long\def\commentw#1{{\ifnum\Comments=1\textcolor{cyan}{\bf --W: #1--}\fi}} 
\long\def\commentc#1{{\ifnum\Comments=1\textcolor{purple}{\bf --C: #1--}\fi}} 
\long\def\commenty#1{{\ifnum\Comments=1\textcolor{blue}{\bf --Y: #1--}\fi}} 
\long\def\changed#1{{\ifnum\Comments=1\textcolor{teal}{#1}\else {#1}\fi}}  
\long\def\changedy#1{{\ifnum\Comments=1\textcolor{blue}{#1}\else {#1}\fi}}  
\long\def\changedforarxiv#1{{\ifnum\Comments=1\textcolor{blue}{#1}\else {#1}\fi}}  
\long\def\changedlastday#1{{\ifnum\Comments=1\textcolor{olive}{#1}\else {#1}\fi}} 
\long\def\changedijcai#1{{\ifnum\Comments=3\textcolor{teal}{#1}\else {#1}\fi}}

\newcommand{\states}{\mathcal{S}}

\newcommand{\actions}{\mathcal{A}}

\newcommand{\transitionfunc}{P}

\newcommand{\rewardfunc}{R}

\newcommand{\horizon}{T}
\newcommand{\policy}{\pi_{\theta}}

\newcommand{\observations}{\mathcal{O}}
\newcommand{\obssets}{\Omega}
\newcommand{\discount}{\gamma}

\begin{document}

\title{Learning a Robust Multiagent Driving Policy for Traffic Congestion Reduction}


\author*[1,2]{\fnm{Yulin} \sur{Zhang}}\email{zhangyl@amazon.com}

\author[2]{\fnm{William} \sur{Macke}}\email{wmacke@cs.utexas.edu}

\author[2]{\fnm{Jiaxun} \sur{Cui}}\email{cuijiaxun@utexas.edu}


\author[3]{\fnm{Daniel} \sur{Urieli}}\email{daniel.urieli@gm.com}

\author[2,4]{\fnm{Peter} \sur{Stone}}\email{pstone@cs.utexas.edu}

\affil*[1]{\orgname{Amazon Robotics}, \orgaddress{\street{300 Riverpark Dr.},
\city{North Reading}, \postcode{01864}, \state{Massachusetts},
\country{United States}}. The work was done prior to joining Amazon}

\affil[2]{\orgdiv{Department of Computer Science}, \orgname{The
University of Texas at Austin}, \orgaddress{\street{2317 Speedway},
\city{Austin}, \postcode{78712}, \state{Texas}, \country{United States}}}

\affil[3]{\orgdiv{General Motors Israel R\&D Labs}}

\affil[4]{\orgdiv{Sony AI}}

\begin{abstract}
\commentc{From Reuth: Complex as a first sentence, maybe start by saying "The advent of automated and autonomous vehicles (AVs) creates opportunities to achieve system level goals using multipe AVs, such as traffic congestion reduction"}

\changedijcai{In most modern cities, traffic congestion is one of the most salient
societal challenges.  Past research has shown that inserting a limited
number of autonomous vehicles (AVs) within the traffic flow, with
driving policies learned specifically for the purpose of reducing
congestion, can significantly improve traffic conditions.  However, to
date these AV policies have generally been evaluated under the same
limited conditions under which they were trained.  On the other hand, to
be considered for practical deployment, they must be robust to a wide
variety of traffic conditions.  This article establishes for the first
time that a multiagent driving policy can be trained in such a way that
it generalizes to different traffic flows, AV penetration, and
road geometries, including on multi-lane roads. \changedyz{Inspired by our successful results in a high-fidelity microsimulation,
this article further contributes a novel extension of the well-known
Cell Transmission Model (CTM) that, unlike past CTMs, is suitable for
modeling congestion in traffic networks, and is thus suitable for
studying congestion-reduction policies such as those considered in this
article.}} 
\commentc{From Reuth: From this abstract the first impression is that it's not clear what's your contribution beyond testing on more complex domains than people did in the past. Can you maybe elaborate in a sentence or two about what's unique in your new policy? Then you can even give up a lot of the discussion on robustness and just say that this new policy is more robust to different traffic condition, which is crucial in real world scenarios.}
\end{abstract}


\maketitle

\section{Introduction}
According to Texas A\&M's  2021 Urban Mobility Report, traffic congestion in 2020 in the U.S.~was responsible for excess fuel consumption of about 1.7 billion gallons, an annual delay of 4.3 billion hours, and a total cost of \$100B~\cite{tamu2021}. A common form of traffic congestion on highways is \emph{stop-and-go waves}, which have been shown in field experiments to emerge when vehicle density exceeds a critical value~\cite{Sugiyama_2008}. Past research has shown that in human-driven traffic, a small fraction of automated or autonomous vehicles (AVs)  executing a controlled multiagent
  driving policy can mitigate stop-and-go waves in simulated and real-world
  scenarios, roughly double the traffic speed, and increase throughput by about 16\%
  ~\cite{stern2018dissipation}.
  Frequently, the highest-performing policies are those learned by deep reinforcement learning (DRL) algorithms, rather than hand-coded or model-based driving policies.
  \commentw{Do we have a citation for this?}
  \commentc{From Reuth: Near-last thing before submission - try to reduce cases where you have 1-2 words on the last line. It gives more space and also looks more professional}
\begin{figure}[t!]
    \centering
    \begin{subfigure}[bt]{0.495\columnwidth}
    \includegraphics[width=1.05\linewidth]{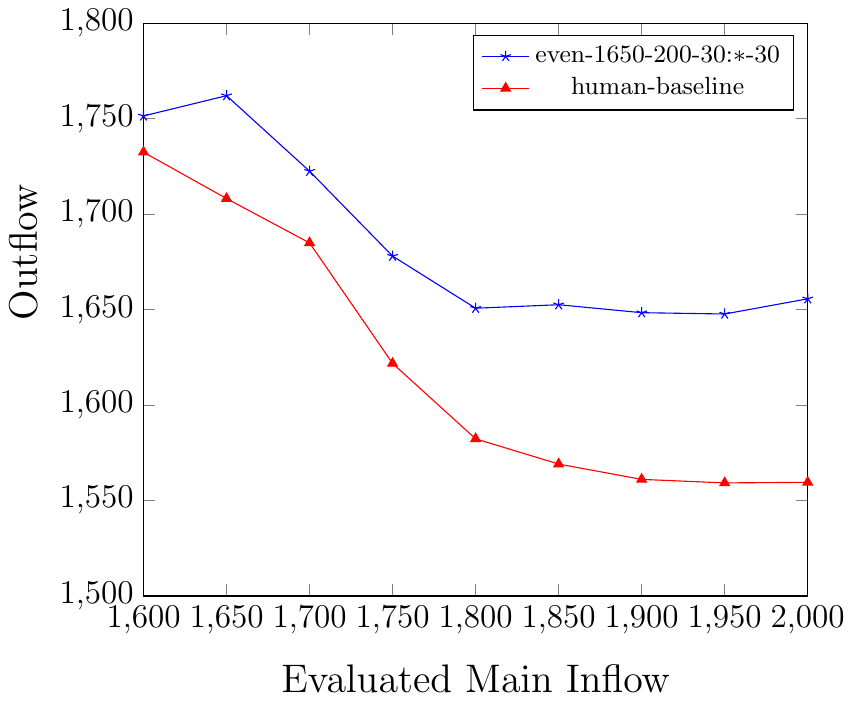}
   \caption{}
    \label{fig:motivation_outflow}
    \end{subfigure}
    \begin{subfigure}[bt]{0.495\columnwidth}
    \includegraphics[width=1.04\linewidth]{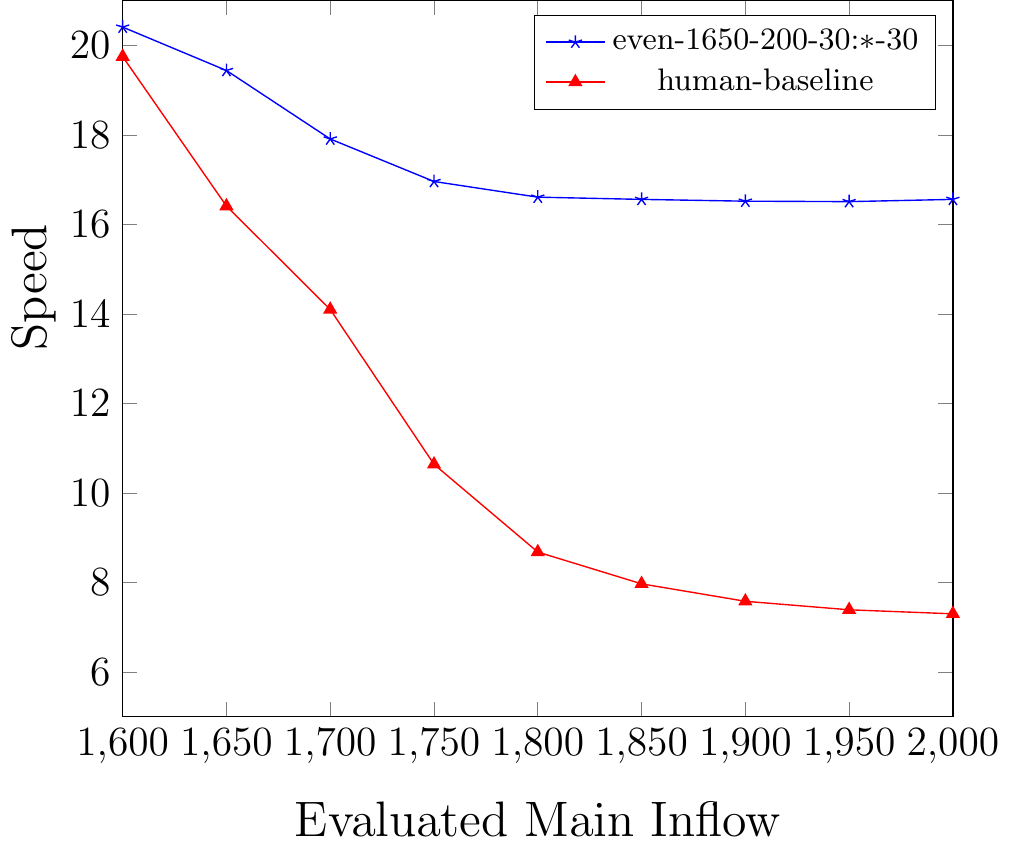}
   \caption{}
    \label{fig:motivation_speed}
    \end{subfigure}
    \begin{subfigure}[bt]{\columnwidth}
    \centering
    \includegraphics[width=0.8\linewidth]{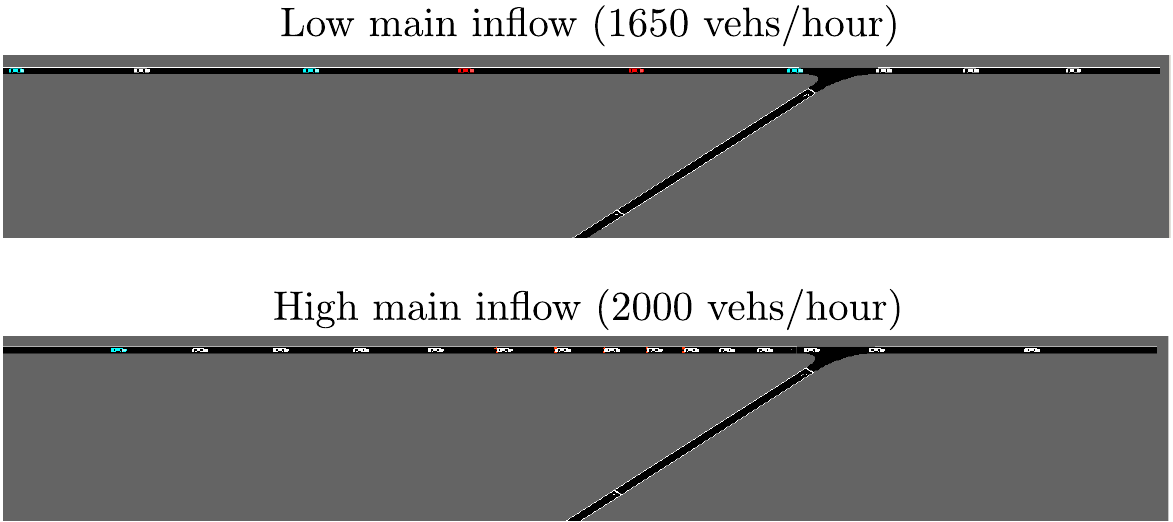}
   \caption{}
    \label{fig:motivation_scenario}
    \end{subfigure}
    \caption{Increasing incoming vehicle flow (the demanded \emph{inflow}) degrades performance of a policy trained with inflow of \SI{1650}{veh/hour}, with respect to both throughput (a) and speed (b). A visual representation (c) is given that shows what this decreased efficiency looks like.  The human baseline shows performance with no AVs in the system (AVP=0).} 
    \label{fig:motivation}
\end{figure}

Any congestion reduction policy executed in the real world will need to perform robustly under a wide variety of traffic conditions such as  traffic flow,  AV penetration  (percentage of AVs in traffic, referred to here as ``AVP"), AV placement in traffic, and road geometry. 
However, existing driving policies have generally been tested in the same conditions they were trained on, and have not been thoroughly tested for robustness to different traffic conditions.
Indeed, their performance can degrade considerably when evaluated outside of the training conditions (Figure~\ref{fig:motivation}). 
Therefore, it remains unclear how to create a robust DRL congestion-reduction driving policy that is practical for real-world deployment.

\commentc{Reuth: Can this be presented as your contribution rather than a challenge that requires *someone* to solve?}
\changedlastday{
\ijcaicommentd{could someone review the story from here  until the end of the intro?}
In this article, we establish for the first time the existence of a robust DRL congestion-reduction driving policy that performs well across  a wide variety of traffic flows, AVP, AV placement in traffic, and several road geometries.}
%
Moreover, we investigate the question of how to come up with such a policy and
what degree of robustness it can achieve. We create a testbed with a diverse,
pre-defined collection of test traffic conditions of real-world interest
including the single-lane merge scenario shown in
Figure~\ref{fig:motivation_scenario}. Such merge scenarios are a common source
of stop-and-go waves on highways~\cite{MitaraiConvective}. 

While there are
different approaches to training robust DRL policies in other domains with
different levels of success, our approach is to systematically search for a
robust policy by varying the training conditions, evaluating the learned policy
on our proposed test set in a single-lane merge scenario, and selecting the
highest performing one.
\changedijcai{The highest performing policy outperforms the human-only baseline with as few as
\SI{1}{\percent} AVs across different traffic conditions in the single-lane
merge scenario.} 

\changedijcai{We further investigate the policy's generalization to more
complex scenarios it has not seen during training, specifically a scenario with two merging
ramps at a variety of distances, and a merge scenario with a double-lane main road, with cars able to change lanes. 
Notwithstanding negative prior results showing that a policy developed in a
single-lane ring road fails to mitigate the congestion on a double-lane ring
road~\cite{cummins2021simulating}, our learned policy outperforms human-only
traffic and effectively mitigates congestion in these more complex scenarios as well.  
}

Inspired by our successful results in a high-fidelity microsimulation,
this article further contributes a novel extension of the well-known
Cell Transmission Model (CTM) that, unlike past CTMs, is suitable for
modeling congestion in traffic networks, and is thus suitable for
studying congestion-reduction policies such as those considered in this
article.
Taken together, this article's contributions and insights take us a step closer towards making the exciting concept of traffic
congestion reduction through AV control a practical reality.

The rest of the article is structured as follows. Section~\ref{sec:related_work} presents related work.
Section~\ref{sec:background} provides a background that includes a formalization of the traffic reduction problem, a description of the DRL setup, and a description of our robustness evaluation conditions.
Section~\ref{sec:robust} describes how the DRL policy is learned and analyzes its empirical performance.
Section~\ref{sec:complex_roads} describes the generalization of our policy  to unseen, complex roads.
Section~\ref{sec:ctm} introduces a novel Cell Transmission Model formulation and use it to empirically characterize the operation of congestion reducing policies.
\changedyzz{Section~\ref{sec:implementation} presents the hyper-parameters used
by the training algorithm and the Cell Transmission Model. The code that generates all data used in this study is available at \url{https://github.com/yulinzhang/MITC-LARG}.}

\commentd{add 1.  link to a placeholder web pages with videos 2. link to code release?}

\commentd{One important thing that is missing in this paper is final  numbers that quantify our robustness (beyond our qualitative conclusions based on visual inspection): by how much we improve throughput and speed upon human baseline (and other tested policies?) how does performance drop as we move away from the training conditions?  and why it's considered "robust"? other? These numbers should appear clearly in both the technical sections, the conclusions section, and perhaps the intro?}
\section{Related work}
\label{sec:related_work}
  
  
 
Traffic optimization has long been a challenging research area with direct real-world impact~\cite{downs2000stuck}. An important research question is how to  mitigate highway \emph{stop-and-go waves},  which have been demonstrated to emerge when vehicle density exceeds a critical value, and to result in reduced throughput and increased driving time~\cite{Sugiyama_2008}. In small-scale field experiments, vehicles controlled autonomously by hand-designed driving policies  successfully dissipated stop-and-go waves, thus reducing congestion~\cite{stern2018dissipation}. The industry-wide development of autonomous vehicles (AVs) 
has inspired researchers to tackle this problem at a larger scale. 

Recent progress in Reinforcement Learning (RL)~\cite{sutton2018reinforcement} has made it possible to learn congestion reduction AV driving policies that perform well in simulation. Using state-of-the-art algorithms,
significant congestion reduction was achieved both in circular roads with a fixed set of vehicles (referred to as \emph{closed} road networks), and  acyclic roads with vehicles entering and leaving the system (referred to as \emph{open} road networks)~\cite{wu2017emergent,kreidieh2018,vinitsky2018}, as compared with simulated human-driven traffic implemented with accepted human driving
  models~\cite{treiber2017intelligent}. 
Most of these past successful driving policies controlled AVs in a \emph{centralized} manner, where a single controller simultaneously processes all available sensing information and sends driving commands to the AVs. More recent efforts focused on developing \emph{decentralized} driving policies which might be harder to learn, but are considered a more realistic option for real-world deployment, as they mostly rely on local sensing and actuation capabilities~\cite{vinitsky2018,cui2021scalable}. \commentc{Reuth: Makes this paper sound incremental - can you distinguish something that you do from what other works have done so far?} this article continues the line of research on decentralized policies but aims to develop one that is robust to real-world traffic conditions of practical interest.

Recent RL techniques for developing robust policies include adversarial training~\cite{pmlr-v70-pinto17a} and domain randomization~\cite{tobin2017domain}. Existing research uses these ideas to  build congestion reduction policies that are robust to some particular traffic conditions. 
Wu et al. present policies that can generalize  on a closed ring road to traffic densities higher and lower than the ones they were trained on, by randomizing densities during training~\cite{9489303flowjournal}. Parvate et al. evaluate the robustness of a \emph{hand-coded} 
controller over different AV penetration and driving aggressiveness~\cite{parvate2020training}. 
This article focuses on learning a driving policy that is robust to different traffic flows, AV penetrations, AV placement within traffic, and road geometries. 

\commentd{Could everyone review the entire following paragraph to make sure that differences are described appropriately?}
\commentc{Quick Pointer: \cite{vinitsky2020optimizing}, Section [Results] -> subsection [B], if anyone would like to check again.}
\changedlastday{In contemporary unpublished 
 work~\cite{vinitsky2020optimizing}, Vinitsky et al. studied a similar setup. In particular, similarly to our work, they developed a robust, decentralized policy that is shared among all AVs for an open road network scenario. On the other hand, our work differs from theirs in several ways. First we focus on merge scenarios, while they focus on bottleneck scenarios. Second, they developed a robust policy by randomizing the training conditions, while we did a systematic sweep of the training conditions to understand how each training condition contributes to the performance of the trained policy. Third, we further examined the robustness of the policy trained from a merge scenario on a more complex road with multiple merging ramps and multiple lanes.}

Finally, to evaluate proposed traffic systems more efficiently, traffic engineers
often make use of more abstract traffic models for their initial
analyses, such as Cell Transmission Models (CTMs)~\cite{daganzo1994cell}.  Unfortunately,
traditional CTMs are not applicable to the topic of this article because
\changedyzz{they do not capture the traffic congestion from multiple merging inflows.}
To alleviate this limitation, in Section~\ref{sec:ctm} we introduce a
novel CTM formulation \changedyzz{that models the traffic congestion by conditionally discounting the merging inflows.}
\section{Background and setup}
\label{sec:background}



\changed{We start by introducing the background and the problem of learning a robust traffic congestion reduction policy.}
\subsection{Road-merge congestion reduction }


Consider a network with a main highway and a merging road, as shown in Figure~\ref{fig:motivation_scenario}. 
There are vehicles joining and leaving the network, and the traffic consists of both human-driven and autonomous vehicles. The human drivers are assumed to be self-interested and optimize their own travel time, 
while autonomous vehicles (AVs) are assumed to be altruistic and have a common goal of reducing traffic congestion. 
Our goal is to come up with a driving policy that controls each AV such that traffic performance is improved.

\commentd{we need to make sure we don't define metrics twice, reconcile with the comment in 3.1} We measure the performance of  policies  in terms of both  \emph{outflow} and \emph{average speed}. Outflow is the number of vehicles per hour exiting the simulation, representing system-level throughput. The average speed represents the time delay it takes an average driver to drive the simulated road. We note that it is important to report both metrics, since scenarios with low and high average speeds could have the same system throughput, such that one is considered congested while the other is not.

A policy can be hand-programmed or learned. Reinforcement learning (RL) has been shown to produce superior policies~\cite{wu2017emergent,kreidieh2018,vinitsky2018}
and is therefore our method of choice.
Congestion reduction driving policies can either be \emph{centralized}, controlling all vehicles simultaneously based on global system information, or \emph{decentralized}, controlling each vehicle independently based on its local observations. Decentralized policies with no vehicle-to-vehicle communication are most realistic, since they mostly rely on local sensing and actuation capabilities~\cite{cui2021scalable,vinitsky2020optimizing}, and are therefore the focus of this article. 


This multiagent traffic congestion reduction problem can be modelled as a discrete-time, finite-horizon 
decentralized partially observable Markov decision process (Dec-POMDP)~\cite{bernstein2002complexity}, denoted as a tuple 
$(\states, \{\actions_i\}, \transitionfunc,\rewardfunc,\{\obssets_i\},\observations, \horizon, \discount)$
 where,
 \changedijcai{
\begin{itemize}
\item $\states$ is a state space representing the location and speed of every vehicle in the network,
\item \changedlastday{$\{\actions_i\}$ is a joint action space for all agents, where \changedyz{$\actions_i\in \mathbb{R}$ is a real number} that specifies an acceleration action for agent $i$, }
\item $\transitionfunc : \states \times \{\actions_i\}\times \states \rightarrow [0,1]$ \changedyz{is a stochastic state transition function, which specifies the probability distribution of target state given the source state and action taken by the vehicle. In this paper, this state transition function is realized via a traffic simulator.}
\item $\rewardfunc : \states \times \{\actions_i\} \rightarrow \mathds{R}$ is a global reward function, 
\changedlastday{\item $\{\obssets_i\}$ is a collection of local observations for each agent (see Section~\ref{sec:DecentralizedDrivingPolicy}), }
\item $\observations : \states \times \{\actions_i\} \times \{\obssets_i
\} \rightarrow [0, 1]$ outputs the probability that each agent receives a specific observation given the next state and the joint action just taken, 
\item $\horizon$ is the episode length, 
\item $\discount \in [0,1]$ is the discount factor of reward.
\end{itemize}
}

\changedijcai{
A decentralized, shared \emph{driving policy} is a probability \changed{density function over the action space}  $\policy : \{\obssets_i\} \times \{\actions_i\} \rightarrow [0,1]$ parameterized by $\theta$ that stochastically maps each agent's local observations to its driving actions. 
}


Throughout this article we use the SUMO traffic
simulator~\cite{krajzewicz2012recent} as the state transition function. SUMO is
a micro simulator that includes accepted human driving models
\cite{krauss1998microscopic,treiber2017intelligent} with  configurable traffic
networks, flows, \changedyz{and driving aggressiveness},
as well as  mechanisms for enforcing traffic rules, safety rules, and basic physical constraints. 
To learn AV driving policies, we use the RLlib
library~\cite{duan2016benchmarking}.
We interface with SUMO and RLlib using UC Berkeley's Flow software~\cite{wu2017flow}. 
\commentp{Somewhere do we describe the default human policy that is used as a baseline?  It doesn't have to be here, but should be introduced somewhere...}



\subsection{RL-based \changedforarxiv{decentralized} driving policy}
\label{sec:DecentralizedDrivingPolicy}
To learn a \changedforarxiv{decentralized} driving policy we use the Proximal Policy Optimization (PPO) algorithm~\cite{Schulman2017Proximal}. 
To facilitate data and computational efficiency and reduce the risk of overfitting, all AVs learn and execute a single, shared driving policy. 
The observation space and reward design used in this article are modeled after
those used by Cui et al.~\cite{cui2021scalable}, which were shown to be effective for decentralized policies.
\commentp{We need to make sure not to make it sound like it's our own past work.  Something like:
The observation space and reward design used in this article is modeled after that used by Cui et al.~\cite{cui2021scalable}:}
The observation for each AV includes
\begin{itemize}
    \item the speed and distance of the closest vehicles in front of and behind it,
    \item the AV's speed,
    \item the AV's distance to the next merging point,
    \item the speed of the next merging vehicle and its distance to the merge junction (\changedforarxiv{assumed to be obtained by the vehicle's cameras/radars, or be computed by some global infrastructure and then shared with all the vehicles}).
\end{itemize}
\commentp{normalized in what way?}

\noindent The reward of the $i$th  AV  at time step $t$ is defined as:
\begin{equation*}
    \label{final_distributed_reward}
    \begin{aligned}
    r_{i,t}=&(1-\mathbb{I}\{done\})
        \bigg(-\eta+(1-\eta)\times\frac{\sum_{j=1}^{n_{t}} v_j}{n_{t}V_{max}}
    \bigg)\\
    & + \mathbb{I}\{done\}\cdot Bonus
    \end{aligned}
\end{equation*}
where $\mathbb{I}\{done\}$ is an indicator function of whether an AV is leaving
the network; $Bonus$ is a constant reward for an AV when it exits the network; the
term $\frac{\sum_{j=1}^{n_{t}} v_j}{n_{t}V_{max}}$ represents the normalized
average speed, where $v_j$ is the speed of vehicle $j$, $n_{t}$ is the total
number of vehicles in the network \changed{at time t}, $V_{\max}$ is the max
possible speed, and $\eta$ is a constant that weights the individual  and the
global reward.  
\commentw{Should we give our value of $\eta$ here?} \commentc{I think it should be removed, removed}

\subsection{Robustness evaluation conditions }
\label{subsec:eval}
\commentp{what do we mean "human drivers"?  We're not acutally using human drivers in our experiments...}
Similarly to past work, our baseline setup  consists of  simulated human-driven vehicles only, \changedyz{where the AVP is 0.} 
In contrast to past work, which typically showed improvement over this baseline in a \emph{single} combination of traffic conditions, our goal is to develop  a robust AV driving policy that improves over this baseline across a \emph{range} of \changed{realistic} traffic conditions, characterized by:
\begin{itemize}
    \item \emph{Main Inflow Rate}: the amount of incoming traffic on the main artery (veh/hour),
    \item \emph{Merge Inflow Rate}: the amount of incoming traffic on the merge road (veh/hour),
    \item \emph{AV Placement}: the place where the AVs appear in the traffic
flow; the AVs can either be distributed evenly or randomly among the simulated
human-driven vehicles.
    \item \emph{AV Penetration}: the percentage of vehicles that are controlled autonomously,
    \item \changed{\emph{Merge road geometry:}  the distance between two merge junctions (in relevant scenarios), and the number of lanes.}
\end{itemize}
\commentd{Are these conditions discretized like the training conditions? if not, we should clarify how is it discretized here}
\changedyz{In this article, we focus on a merge inflow rate of \SI{200}{veh/hour} and a main inflow rate in the range of [1600, 2000] {veh/hour} since these values tend to lead to congestion in the baseline (AVP=0) conditions. We vary all the other parameters as follows: AV penetration (AVP) is set to be within [0, 40] percent to represent a realistic amount of controllable AVs that can be expected in the coming years, and the placement of the AVs can either be random or even.}
\ijcaicommentd{why small is good?} 
For \emph{even
placement}, AVs are placed every N human-driven vehicles in a lane.
For \emph{random placement}, AVs are placed randomly among simulated human-driven
vehicles. Merge road geometries include one or two merges at distances that vary
between [200, 800] meters, \changedijcai{and the main road can have one or two
lanes}.

\commentd{we need to make sure that anywhere we mention Outflow as our metric, will now mention Outflow and speed as our metrics}

\section{\changedijcai{Learning a robust policy in the single-lane merge scenario}}
\label{sec:robust}
While real-world congestion-reducing driving policies need to operate
effectively in a wide variety of traffic conditions, most past research has
tested learned policies under the same conditions on which they were trained. 
\commentp{One approach to this issue is learning to deal with distribution
shift.  but we're not dealing with that - we're just trying to find a single
robust policy.  We should probably mention this in related work as an
alternative approach.} 
Since in the real world it is impractical to deploy a separate policy for each
combination of conditions, our primary goal is to understand whether it is
feasible to learn a \emph{single} driving policy that is robust to real-world
variations in traffic conditions. 


\ijcaicommentp{Actually, I think it's a tradeoff.  Less diverse should lead to better performance in the specific scenarios the training data comes from - because the policy doesn't have to cover as many conditions.
I suggests phrasing this passage as more of a hypothesis to be tested.  We hypothesize that training under the following conditions will lead to a policy that generalizes across all the conditions we're interested in...}
\changedijcai{The performance of an RL-based driving policy depends on the traffic conditions under which it is trained. We hypothesize
that the policy trained under high inflow, medium AV penetration, and random vehicle placement is robust in a range of traffic conditions defined in Section~\ref{subsec:eval} for a single-lane merge scenario.
We test this hypothesis by comparing 30 policies, each of which is trained under a combination of traffic conditions specified below in
Section~\ref{subsec:discretization}. The training of each policy takes about 7 hours on a \SI{3.7}{\GHz} Intel 12 Core i7 processor. \changedyz{SUMO has built-in stochasticity which includes vehicle departure times and vehicle driving dynamics.} Hence, each policy, including human-only baseline, is evaluated 100 times using a fixed set of 100 random seeds, and each evaluation takes about one hour. 
After we identify a policy that generalizes well across traffic conditions in the training road geometry, a later section will describe an  evaluation this policy on more complex road geometries unseen at training time.
}\ijcaicommentd{I think we decided to modify this story in our last meeting, so that it's not a hypothesis, but creating a test set, varying the training conditions, and selecting the best policy through analyzing the plots (or something like that?). That would also connect well to 4.1 which starts from a point that we actually change training traffic conditions}

\subsection{Discretization of traffic conditions \changedijcai{for training}}
\label{subsec:discretization}

\changed{Since there is an innumerable set of possible traffic conditions, for the purpose of training we discretize traffic conditions along their defining dimensions to a total of 30 representative combinations of conditions, as follows.
We consider main inflows of 1650, 1850, and 2000 \SI{}{veh/hour} which result in low, medium, and high congestion. We discretize AV placement in traffic to be random or even-spaced.
Finally, we  discretize the training AV penetration into 5 
levels: \SI{10}{\percent}, \SI{30}{\percent}, \SI{50}{\percent},
\SI{80}{\percent}, \SI{100}{\percent}. Based on this $3\times2\times5$
discretization, we train 30 policies, \ijcaicommentd{I would remove what's in the parenthesis since currently there is no proposed policy} one for
each combination.}

\changed{Each trained policy is then evaluated across the range of  traffic
conditions described in Section~\ref{subsec:eval}, leading to two performance
values (outflow and average speed) on each testing condition for each policy. 
We plot these results using the following convention. 
The label of a data point consists of two parts:
(i) the training conditions of the policy to be evaluated, and (ii) the policy's evaluation conditions. 
The policy's training conditions indicate the vehicle placement,
main inflow, merge inflow, and AVP, separated by ``-". For example, ``random-2000-200-30" denotes the policy trained under random vehicle placement with main inflow \SI{2000}{veh/hour}, merging inflow \SI{200}{veh/hour}, and \SI{30}{\percent} AVP.}
\changedy{
The evaluation conditions also consist of vehicle placement, main inflow, merging inflow, and AVP. In this article, the merging inflow is always fixed to be \SI{200}{veh/hour} and the vehicle placement is specified separately from the graph label. 
Therefore we only  specify the evaluation-time main inflow and AVP to indicate the evaluation condition for each data point. Hence, each evaluation result is labeled as a 6-tuple, where the first four elements describe the training conditions and the remaining two describe the evaluation conditions. For example, ``random-2000-200-30:1800-10" labels the result of policy ``random-2000-200-30" evaluated under main inflow \SI{1800}{veh/hour} and AVP \SI{10}{\percent}. We further use ``\textasteriskcentered{}" in the evaluation condition to denote which evaluation condition varies in a plot. For example, ``random-2000-200-30:1800-\textasteriskcentered{}" indicates that the policy ``random-2000-200-30" was evaluated under main inflow of 1800 and varying AVPs; ``random-2000-200-30:\textasteriskcentered{}-10" indicates that policy ``random-2000-200-30" was evaluated under AVP \SI{10}{\percent} and varying main inflows.} 

\subsection{\changedijcai{Robustness to vehicle placement, AV penetration and inflow}}
\begin{figure*}[ht]
    \centering
    \begin{subfigure}[!bt]{0.99\columnwidth}
    \includegraphics[width=\linewidth]{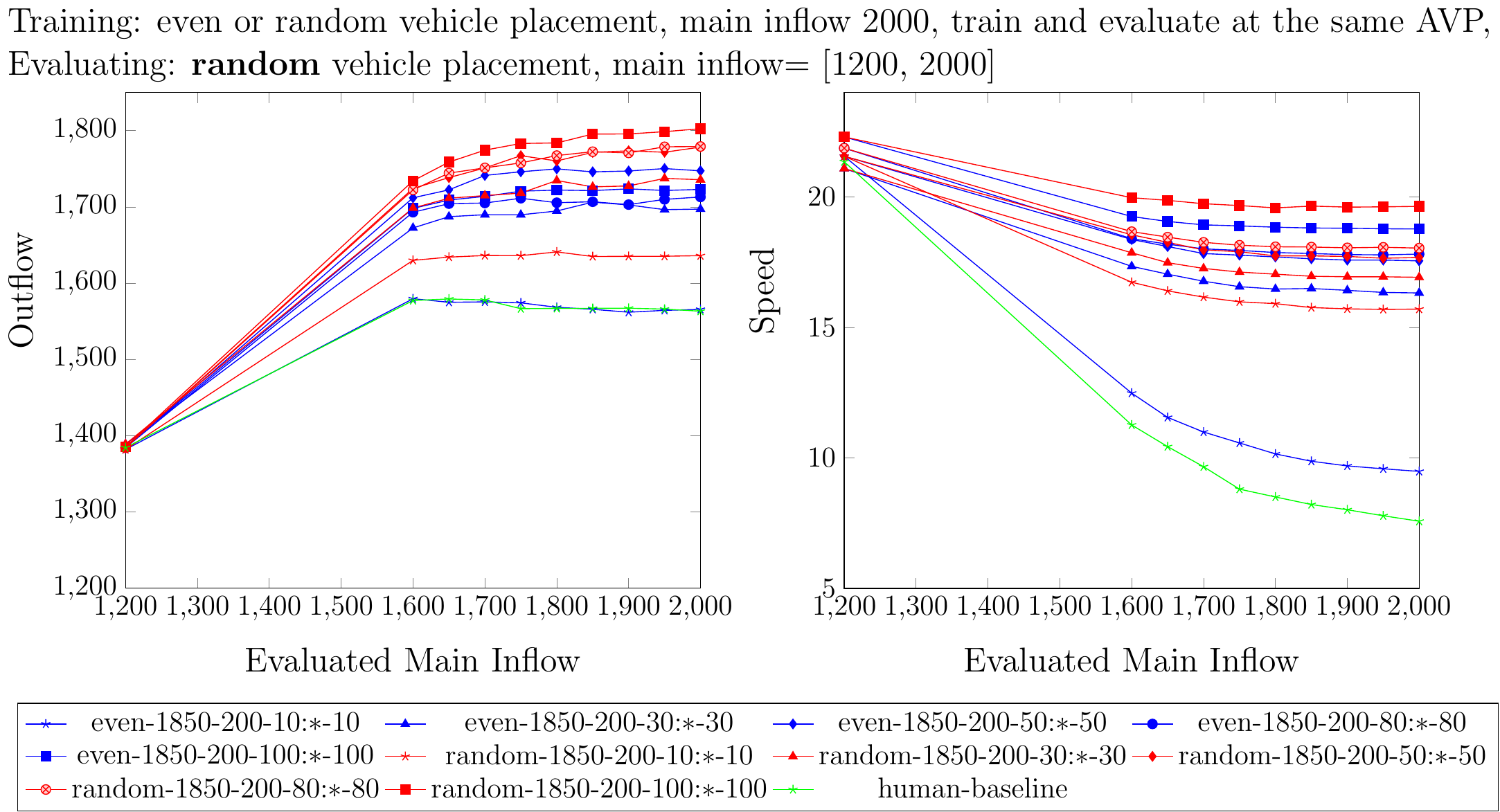}
	\caption{}
\label{fig:veh_placement_rand_eval}
    \end{subfigure}
    \hfill 
    \begin{subfigure}[!bt]{0.99\columnwidth}
    \includegraphics[width=\linewidth]{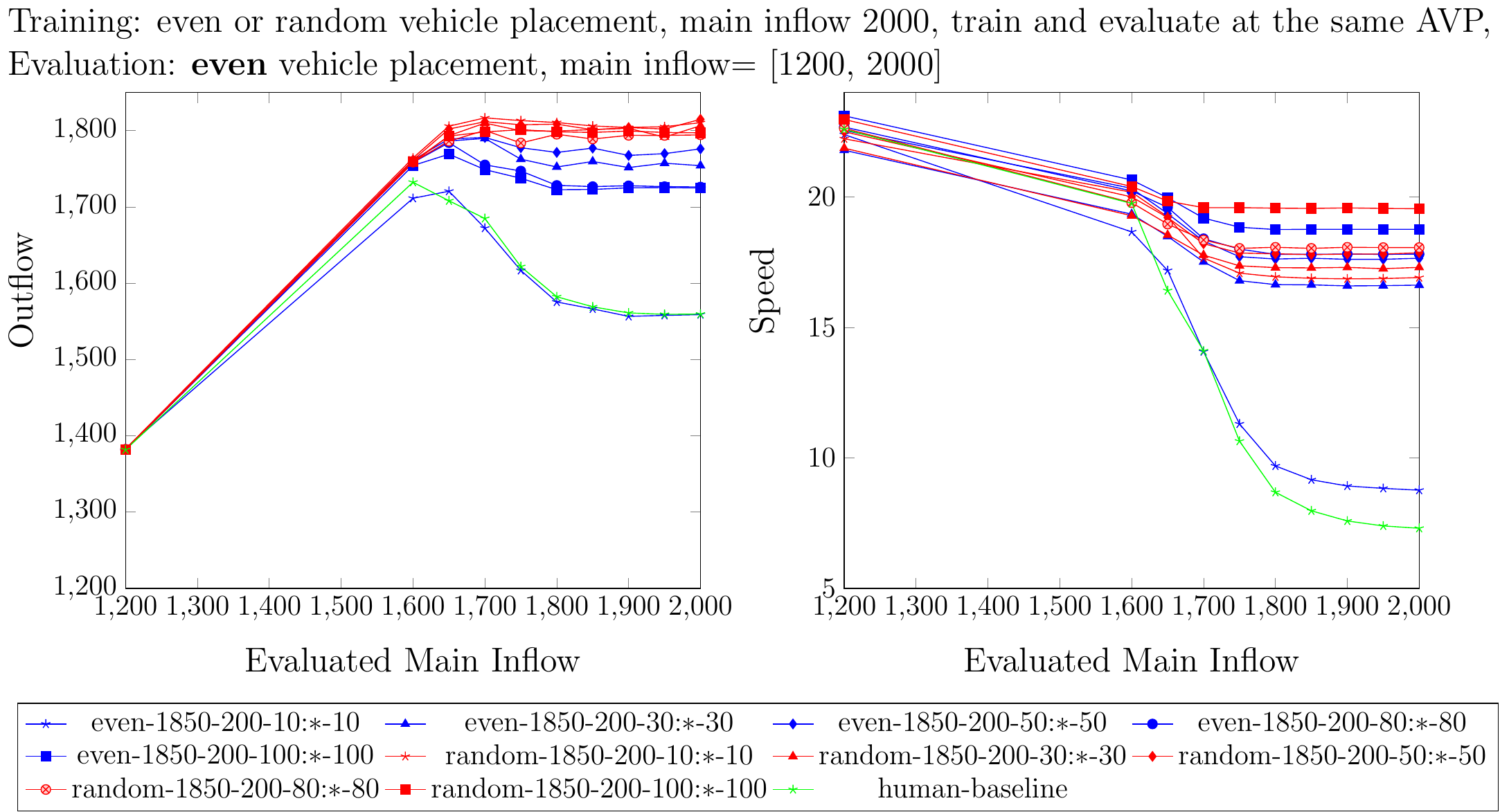}
	\caption{}
    \label{fig:veh_placement_even_eval}
    \end{subfigure}
    \begin{subfigure}[!bt]{0.49\textwidth}
    \includegraphics[width=\linewidth]{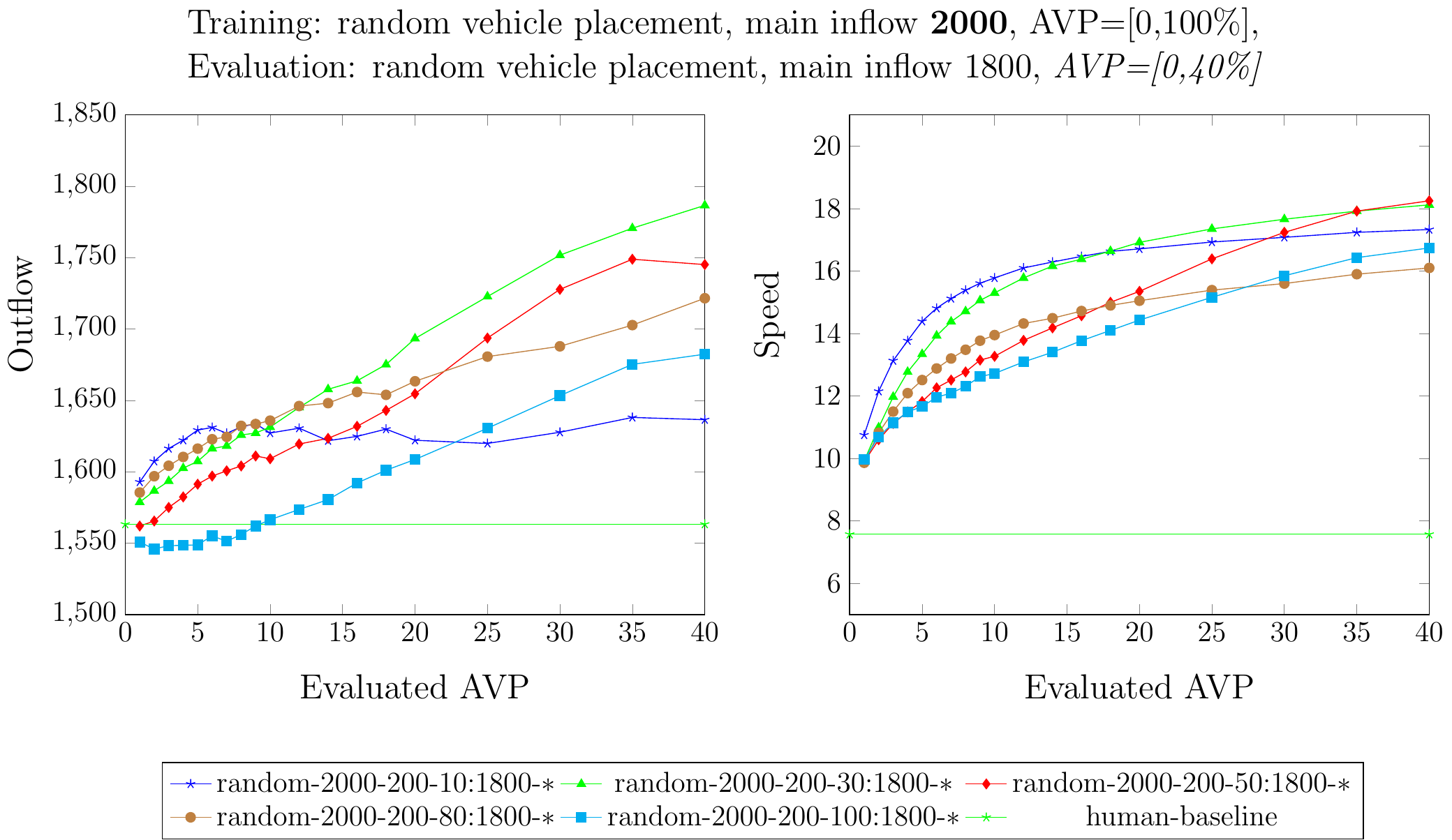}
    \caption{}
    \label{fig:best_avp_avp_2000}
    \end{subfigure}
    \hfill
\begin{subfigure}[!bt]{0.49\textwidth}
    \includegraphics[width=\linewidth]{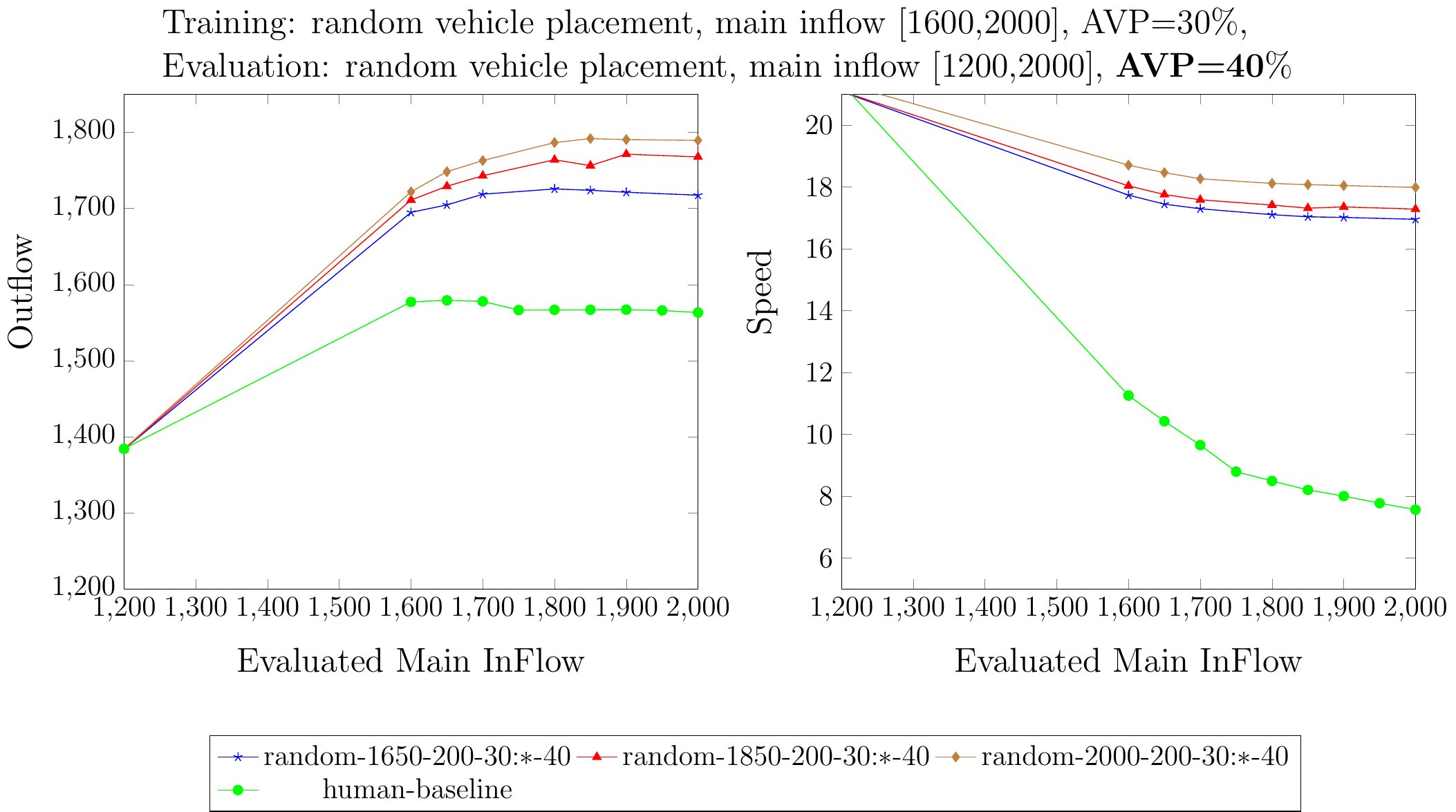}
    \caption{}
    \label{fig:best_inflow_vs_inflow_40}
    \end{subfigure}
    \caption{
	Results of policies trained under different AV placements, AV
penetrations, and main inflows. Figure (a)--(b): we show that  the policies trained under random
vehicle placement outperform their counterparts trained with even placement, when evaluated under both random and even vehicle placement. Figure (c): we fix the evaluation inflow at a medium level and
find that a training AVP of \SI{30}{\percent} is the most robust when varying evaluation AVPs; Figure
(d): we fix the evaluation AVP, and verify that main inflow \SI{2000}{veh/hour} is  the most robust 
when varying evaluation inflows. 
} 
    \label{fig:experimental_results}
\end{figure*}
\label{subsec:vehicle_placement}
In this section, we test our hypothesis that training with high inflow,
medium AV penetration, and random vehicle placement yields a robust policy, by showing representative slices of the evaluation results. 

\changedijcai{
We start by showing that the policies trained under random vehicle placement
outperform policies that are trained under even vehicle placement.
\ijcaicommentd{$\leftarrow$ That's the sequential process I think we could avoid
and save space, saying directly that inspection of performance plots showed that
random-2000-200-30 is a top performer over the entire test set, and showing
slices of it vs. other policies, by varying both inflows and AVP in the x-axes.
Same for the next paragraphs - I think we can avoid them and condense everything
to the way suggested here}
The performance of a representative subset of these policies is depicted in
Figure~\ref{fig:veh_placement_rand_eval} and~\ref{fig:veh_placement_even_eval}.}
The red curves represent the evaluation results for the policies trained under
random vehicle placement, and the blue curves represent the results for the
policies trained under even vehicle placement. These policies are evaluated
\changed{using the outflow and average speed metrics} under both random vehicle
placement (Figure~\ref{fig:veh_placement_rand_eval}) and even vehicle placement
(Figure~\ref{fig:veh_placement_even_eval}). 
\changed{When evaluating on either random placement or even placement, the
policies trained with random placement outperform the human baseline as well as their counterparts trained with even
placement.} 
\changed{Specifically, the results in} Figure~\ref{fig:veh_placement_rand_eval}
\changed{confirm the intuition} that when evaluated \changed{with} random
vehicle placement, the policies trained under random vehicle placement
\changed{should} have better performance than their counterparts trained
\changed{with} even vehicle placement. \changed{However, counter-intuitively},
random placement \changed{at} training \changed{time also results in more robust
policies when testing under \emph{even} placement.} 
\changed{We hypothesize that this performance increase is due to the more
diverse data collected when RL vehicles are randomly placed.} 

\changedijcai{
Next, we confirm the intuition that the polices trained under medium AV penetration are better than others.} Figure~\ref{fig:best_avp_avp_2000} 
show when fixing the main inflow, the policies trained under AVP \SI{30}{\percent} (green curve with triangle) are competitive in both their outflow and average speed when evaluated under varying AVPs. They have the best performance across a large range of the evaluation AVPs. 
We hypothesize that these mid-range AVP values during training perform
best since (i) lower AVP may not encounter enough situations with densely distributed AVs, and (ii) higher AVP may not encounter enough situations with sparsely distributed AVs.

\changedijcai{
Finally, we test the hypothesis that the policies trained under high inflow are robust. 
}
When fixing the AVP and varying main inflow during evaluation,
Figure~\ref{fig:best_inflow_vs_inflow_40} shows that the policy trained
under main inflow \SI{2000}{veh/hour} (brown curve) 
has better performance 
than policies trained with different main
inflows, in terms of
both outflow and average speed. 
We hypothesize that the policies trained under the highest inflow
\changed{outperform others because a higher main inflow yields more diverse
vehicle\commentd{AV?} densities at training time. Specifically, the simulation
dynamics can lead high inflow to include both dense and sparse vehicle\commentd{AV?} placement, while a lower main
inflow tends to mostly result in  a sparse vehicle distribution.
}

\changedijcai{Verifying our hypothesis, we find that the policy
``random-2000-200-30", which is trained under random vehicle placement, main
inflow \SI{2000}{veh/hour}, merge inflow \SI{200}{veh/hour}, and AVP
\SI{30}{\percent}, outperforms the alternatives in terms of robustness. In the single-lane merge
scenario, this policy achieves significant improvement over the human-only
baseline when the AVP is greater than or equal to \SI{1}{\percent} during
deployment (with p-value 0.05 as the cutoff for significance).}

\section{\changedijcai{Deploying the learned policy to more complex roads}}
\label{sec:complex_roads}
\changedijcai{We learned a robust policy in a single-lane merge scenario. To
push this policy one step further toward a real-world deployment, we test this
policy's robustness to more complex road structures: roads with two merging
ramps, and double-lane roads. 
}

\subsection{\changedijcai{Deployed to roads with two merging ramps}}
\label{subsec:doublemerge} 
\begin{figure*}[hbt]
    \centering
    \includegraphics[width=\linewidth]{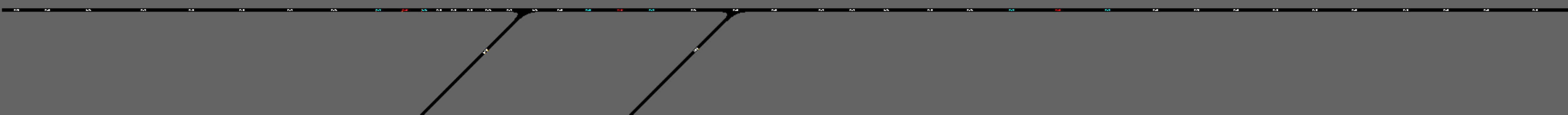}
    \caption{A merge road with two merging on-ramps.}
    \label{fig:ramps-merge}
\end{figure*}

\commentp{I don't think we can call the 2 merges "complex".  Let's stick with "more complex"}
 \changedijcai{We first deploy the selected policy on more complex road
structures, which have two merging roads at varying distances as shown in Figure~\ref{fig:ramps-merge}, and evaluate the performance of the learned policy with respect to the distance between these two ramps.} 

Consider the merge scenario with two \changed{merging ramps}:
\commentd{why is figure 5 here while figure 6 is referred to from the previous section?}\commentc{For a compact organization of figures to fit in the 8-page requirement, we have to make such order switch of figures }
\changed{the first merging ramp is located 500 meters from the simulated main road's start, the second merging ramp is located 200, 400, 600, or 800 meters after the first, the total length of the main road is 1500 meters, and the total length of the merging roads is 250 meters.} 
We \changed{tested the random-2000-200-30} policy with random \changed{AV placement,} main inflow \changed{of} \SI{1800}{veh/hour}, merge inflow \SI{200}{veh/hour}, across a range of AV penetrations \changed{and the above gaps between the two merging roads}. 

\begin{figure}[ht!]
    \centering
    \includegraphics[width=\linewidth]{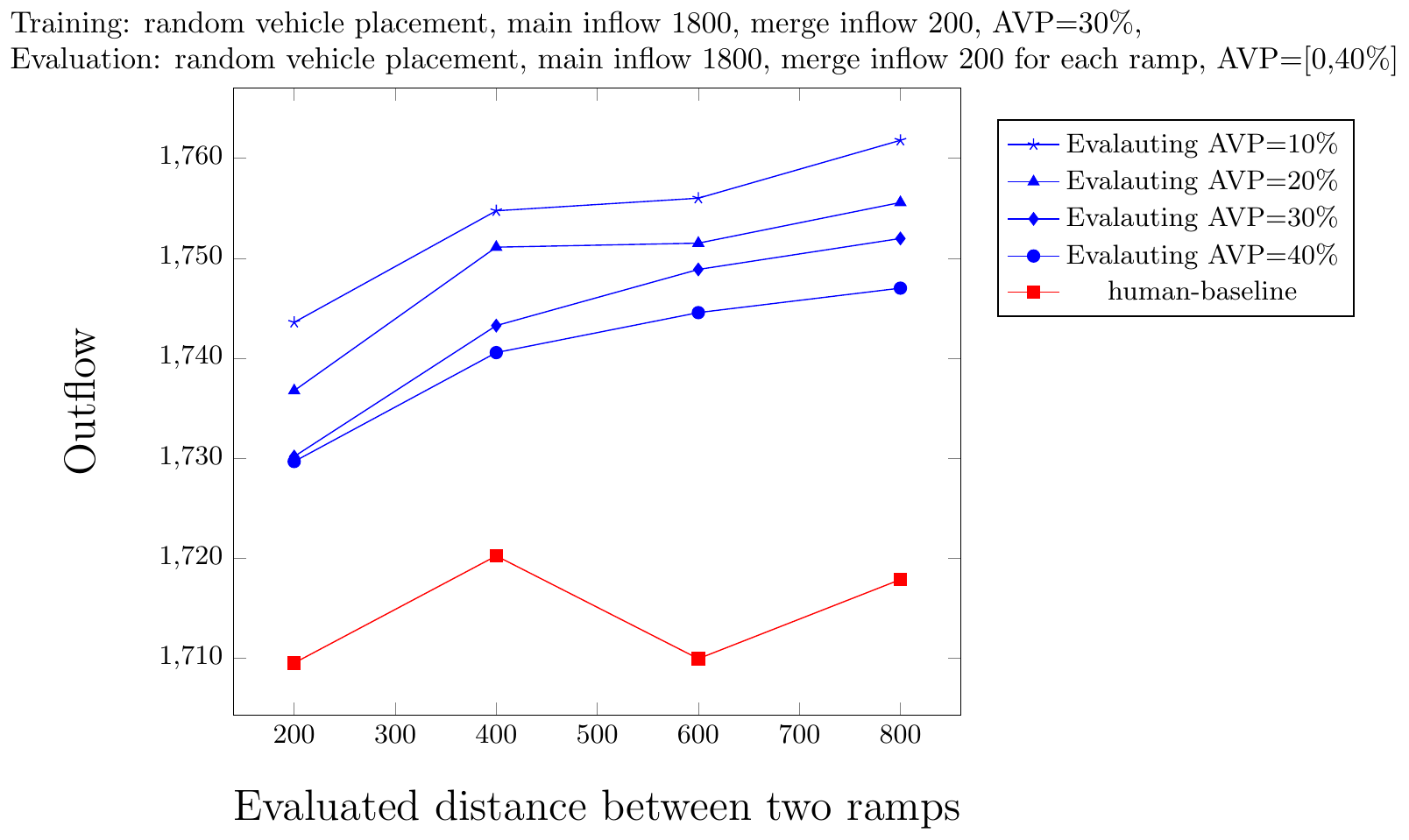}
    \caption{Results of deploying the selected training policy on roads with two on-ramps.\commentd{fix: training flow was 2000}}
    \label{fig:ramps}
\end{figure}

The results are shown in Figure~\ref{fig:ramps}, where the \changed{blue curves show the} performance of the policy to be tested with different AVP values, and the \changed{red curve shows the human baseline's} performance.
The \changed{random-2000-200-30} policy is best 
when the distance between the two on-ramps is large. As we decrease this distance, the performance gap from the human baseline decreases, but remains positive even when the merging ramps are just 200 meters apart, which is the setup that is most different than the training conditions, as explained next. 
When the distance between on-ramps is small,  the traffic congestion at the second merging ramp interferes with the traffic flow at the first merging ramp, but is not observable to the RL vehicles approaching the first ramp. 
\changedlastday{As we increase the distance between these two merging ramps, such interference decreases and the traffic flow approaching these two merging ramps can be treated by the AVs increasingly independently. As a consequence, when these two merging ramps become further away from each other, the decision making processes for the AVs become similar to those on the single-lane merge roads --- they only need to consider the traffic flow at the next incoming junction. To summarize, the selected policy slightly reduces traffic congestion in the two-ramp scenario; and its performance improves as the distance between these two ramps increases.} 

\subsection{\changedijcai{Deployed to double-lane merge roads}}
\begin{figure}[ht!]
    \centering
    \includegraphics[width=\linewidth]{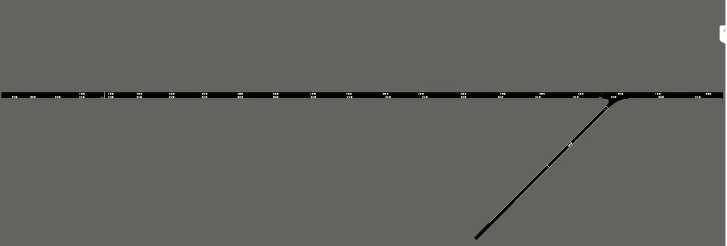}
    \caption{A double-lane merge scenario.}
    \label{fig:double_lane_road}
\end{figure}

Urban highways often consist of multiple lanes. Thus past research
suggesting that AVs might \emph{increase} traffic congestion on multi-lane
roads~\cite{cummins2021simulating} has (rightfully) raised concerns about the
practical deployability of systems like the one considered in this article.
Contrary to those results, we find that AVs can reduce congestion even in
multi-lane scenarios.  Specifically, we consider a double-lane merge road, by
adding a second lane in the main road, as shown in Figure~\ref{fig:double_lane_road}. 
Similarly to the single-lane merge
scenario, the vehicles in the right lane must yield to the
vehicles from the merging lane and may cause potential congestion in the right
lane. But the vehicles in the left lane have the right of way when passing the
junction. As a consequence, the vehicles in the left lane tend to move at a
faster speed, and there will be more vehicles changing from right to left for
speed gain than the number of vehicles changing from left to right.
Those lane-changing vehicles cause additional stop-and-go waves in
the left lane. 

\begin{figure}[ht]
    \centering
    \begin{subfigure}[!bt]{0.99\columnwidth}
    \includegraphics[width=\linewidth]{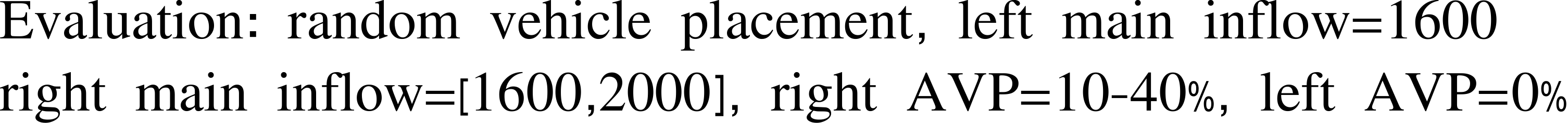}
    \end{subfigure} \\
    \begin{subfigure}[!bt]{0.49\columnwidth}
    \includegraphics[width=\linewidth]{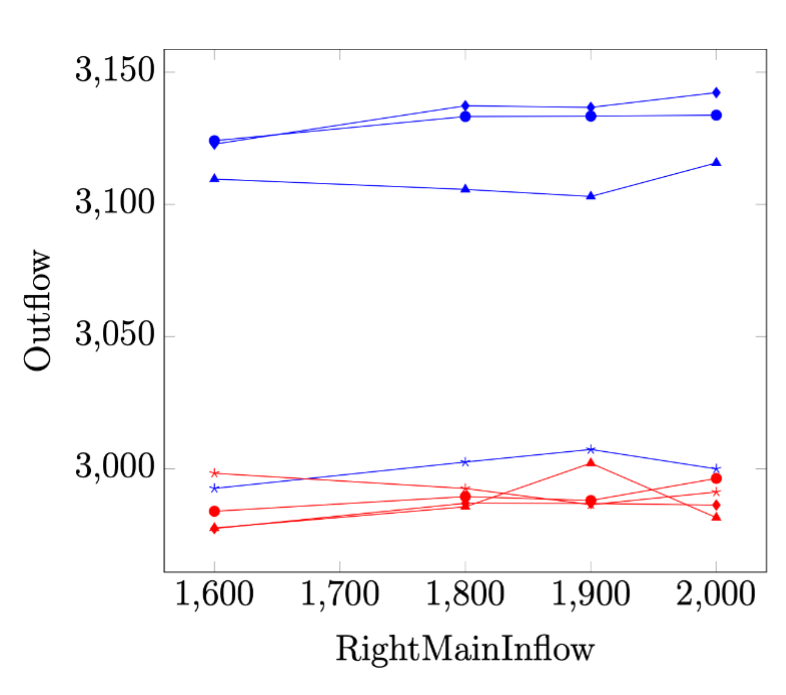}
    \end{subfigure}
    \hfill 
    \begin{subfigure}[!bt]{0.49\columnwidth}
    \includegraphics[width=\linewidth]{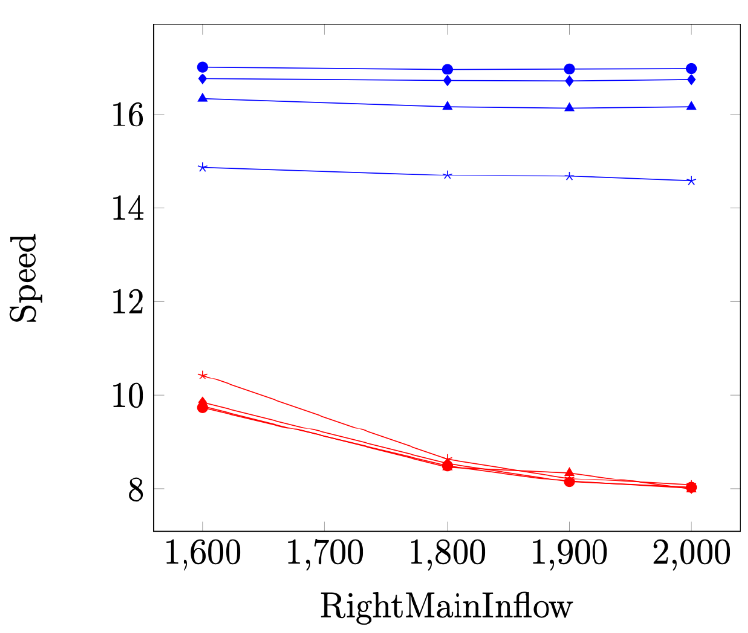}
    \end{subfigure} \\
    \begin{subfigure}[!bt]{0.99\columnwidth}
    \includegraphics[width=\linewidth]{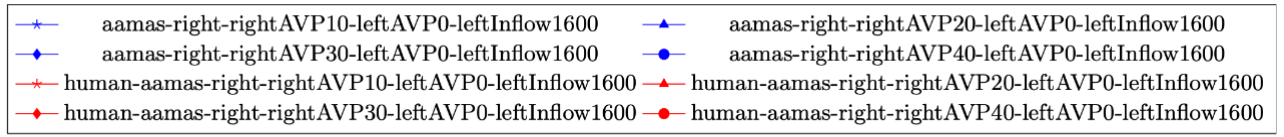}
    \end{subfigure}
    \caption{
    Results of deploying the selected training policy on the double-lane merge roads. }
    \label{fig:double-lane}
\end{figure}

We test the robustness of our selected policy when deployed in the right lane in this new road
structure. In our experiments, 
the left lane contains no AVs and an inflow of  \SI{1600}{veh/hour}  human-driven vehicles, and the right lane contains an AVP of \SI{10}{\percent}--\SI{40}{\percent} that are controlled by our selected policy.
Figure~\ref{fig:double-lane} shows that for right main inflows of \SI{1600}-\SI{2000}{veh/hour}, our policy improves outflow by about \SI{4}{\percent} and traffic speed by about 2x compared with human-only traffic.
We observed 
that the learned policy, mitigating the congestion in the
right lane b also reduces the amount of lane-changing vehicles since the right
lane is less congested.  Hence, the policy trained on the single-lane merge road
generalizes well in the double-lane merge scenario.

\section{Abstract Analysis in an Extended Cell Transmission Model}
\label{sec:ctm}
\changedyz{
The results described in Sections~\ref{sec:robust}
and~\ref{sec:complex_roads} demonstrate for the first time a driving
policy that generalizes over a variety of traffic conditions and road
structures of real-world interest, and therefore takes us closer to a
practical reality of traffic congestion reduction through AV control.
This demonstration was based on extensive empirical comparisons in SUMO,
a high-fidelity microsimulator of traffic, that required hours of computation on a \SI{3.7}{\hertz} CPU.}

As summarized in Section~\ref{sec:related_work}, traffic engineers
often use abstract traffic models, such as such as Cell Transmission Models (CTMs)~\cite{daganzo1994cell} to prototype new traffic protocols.
However, since
traditional CTMs are not equipped to model traffic congestion from multiple merging inflows, 
they are not directly useful for our topic of study.  
In this section, we therefore introduce a
novel CTM formulation \changedyzz{that models the traffic congestion by conditionally discounting the merging inflows.}

\changedyz{We validate that this new abstract formulation
accurately reflects microsimulation results, and then we empirically
characterize the operation of our proposed congestion-reducing policies
in CTM.  The benefit of such a characterization is two-fold.  First, it
provides insights into the connection between the local and global
impact of such policies on traffic, thus contributing to the
explainability of current and future driving policies.  Second, these
insights can guide the development of robust AV policies for large-scale
multilane scenarios that are too slow to exhaustively explore using
micro-simulations such as SUMO.}

Our analysis proceeds according to the following steps:

%
\begin{itemize}
    \item Discretizing the road into basic segments (referred to hereby as \emph{cells})
    \item Empirically fitting a fundamental diagram of traffic flow for each cell. 
    \item Using these fundamental diagrams to construct a novel extension of a CTM for the merge scenario in Figure~\ref{fig:motivation}.
    \item Validating this CTM against SUMO by showing that their global behaviors  (overall simulation inflow and outflow) are similar.
    \item Further introducing a novel extension of CTM to model the double-lane merge scenario from Figure~\ref{fig:double_lane_road}, and similarly  validating its global behavior against SUMO's.
    \item Using these CTMs to extract insights regarding the desired local (intra-cell/segment) behavior of a policy to improve global traffic flow (simulation outflow), which in turn provides a direction for designing congestion-reduction policies for large-scale multilane scenarios that are too slow to explore by exhaustive simulations.
\end{itemize}


\subsection{Discretizing road into cells and fitting their fundamental diagrams} 
\begin{figure}[ht!]
    \centering
    \includegraphics[width=1\linewidth]{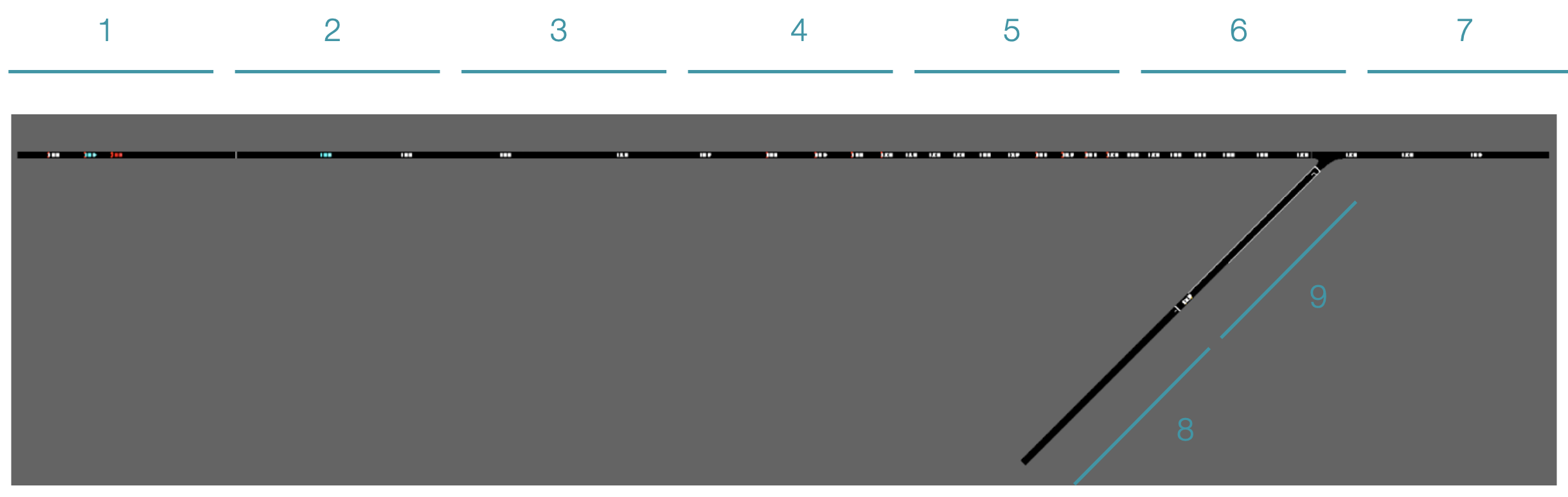}
    \caption{Discretizing the road into cells.}
    \label{fig:road_discretization}
\end{figure}

We start by discretizing the single-lane merge scenario from Figure~\ref{fig:motivation} into \SI{100}{\metre} cells, as shown in Figure~\ref{fig:road_discretization}. \changedyz{The cell length of \SI{100}{\metre} was selected to be small enough to capture the local traffic around each autonomous vehicle, and  large enough for computational efficiency.}

\begin{figure}[ht!]
\begin{subfigure}[bt]{\columnwidth}
    \centering
    \includegraphics[width=0.8\linewidth]{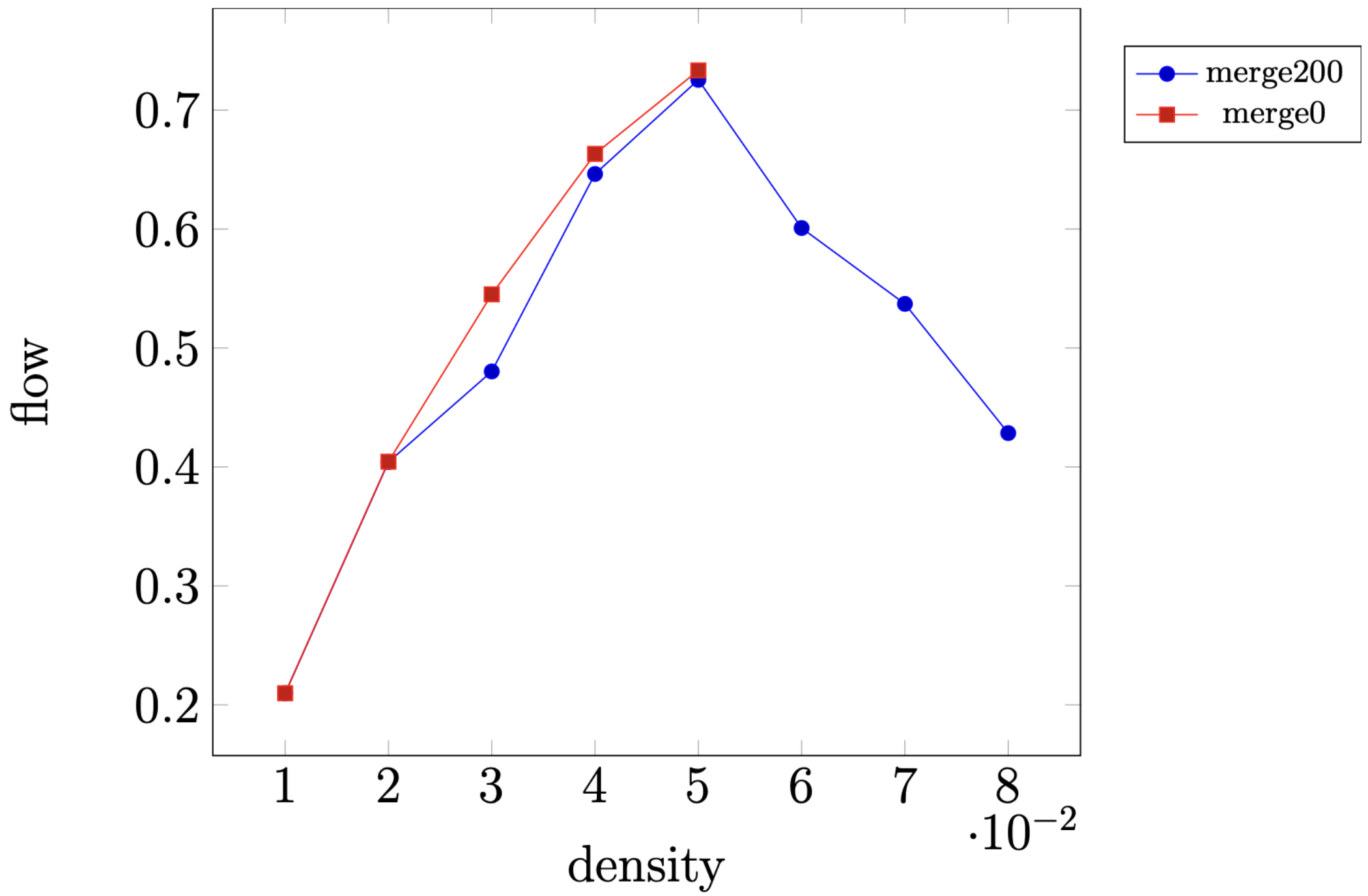}
    \caption{\label{fig:fundamental_diagram}}
\end{subfigure}
\begin{subfigure}[bt]{\columnwidth}
    \centering
    \includegraphics[width=0.8\linewidth]{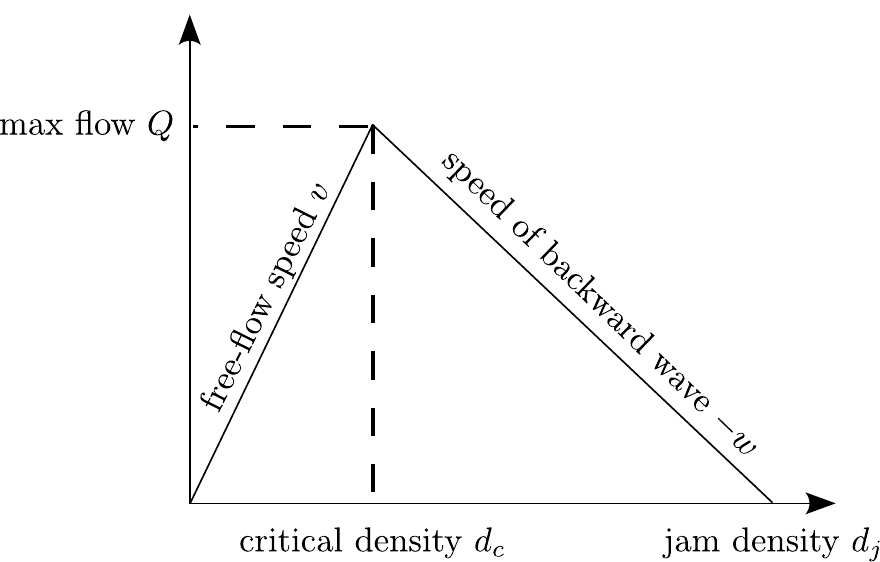}
    \caption{\label{fig:triangular-fd}}
\end{subfigure}
\caption{(a). Profiling the flow density relation of a cell in SUMO: the x axis is the density (veh/m), and the y axis is the flow (veh/s).  (b). A triangular fundamental diagram and its parameters.}
    \label{fig:fd}
\end{figure}

Next, we import from traffic flow theory the concept of a traffic fundamental diagrams, which yields the relationship between the traffic density and traffic flow~\cite{geroliminis2008existence}. 
To obtain a fundamental diagram for each cell in SUMO, we profiled the instantaneous density and average speed, and calculated the flow as the product of instantaneous density and average speed. Since the fundamental diagram characterizes the intrinsic properties of the road conditions (such as capacity and speed limit), the diagram is independent of the  inflows. In Figure~\ref{fig:fundamental_diagram}, we profile the fundamental diagram of merge inflow 200 (blue) and 0 (red) veh/hour. For merge inflow 0, there is no congestion in the road and so the density of the cells will never be higher than 0.05 veh/second. From this fundamental diagram, we can observe that the results for both of these merge inflows are almost the same. Similarly, we observe the same fundamental diagrams for all cells, and therefore we model every cell with the same fundamental diagram. 

Based on the observed data, we see that the fundamental diagram is close to a triangular shape. Hence, we fit a triangular fundamental diagram as shown in Figure~\ref{fig:triangular-fd}, which is defined by the slope before the peak (called free-flow speed $v$), maximum flow $Q$ and its corresponding density (critical density $d_c$), slope after the peak (speed of the backward wave $w$), and the density to reach 0 flow (jam density $d_j$).


\subsection{Constructing an extended CTM from fundamental diagrams.}
Next, we introduce an extended CTM, which models a single-lane merge scenario using the fitted fundamental diagram as a model of intra-cell behaviors.
We start by defining two additional parameters that characterize all cells: 
\begin{itemize}
    \item $Q =d_c \times v$ is the maximum number of vehicles that can flow into a cell when the clock advances,
    \item $N =100\times d_j$ is the maximum number of vehicles in a cell, where $100$ is the cell length.
\end{itemize}
Let $y_i(t)$  and $n_i(t)$ be the inflow and number of vehicles in cell $i$ at time $t$. The inflow is upper bounded by the total number of vehicles in the upstream cells, maximum number of vehicles that can flow into the current cell, and the number of available positions in the cell discounted by the  ratio of wave and free-flow speeds~\cite{tna2022stephen} 
i.e.,
$$
y_i(t) = \min \big\{n_{i-1}(t), Q, \frac{w}{v} [N-n_i(t)]\big\}
$$
When the merge traffic exceeds a certain threshold, more vehicles on the main road will have to slow down or stop to yield to merging traffic. This causes a reduction in the inflow right after the junction, i.e., at cell $7$. To model this, we introduce a conditional penalty factor $\alpha$ to discount the inflow of the cell after the merge: if the flow from the merge road is larger than some threshold $\beta$, then the inflow of the downstream cell is discounted by $\alpha$, i.e., 
$$y'_7(t)=\alpha \times y_7(t),$$
where both $\alpha$ and $\beta$ are hyper-parameters.

Using the above rules, we can update the number of vehicles at cell $i$ at time $t+1$ by adding the inflow and subtracting the outflow at time $t$:
\begin{equation}
n_i(t+1) = n_i(t) + y_i(t) - y_{i+1}(t)
\label{eq:ctm}
\end{equation}
The scenario's overall inflow and outflow are then the inflow of the left most cell (cell 1) and outflow of the right most cell (cell 7). The video of the CTM simulation for single-lane merge scenario can be found here: \url{https://youtu.be/tcAP-adSatM}.
\commentdu{it would be good to put here links to the CTM videos that appear in slide 9}

\subsection{Validating the single-lane CTM against SUMO}
\begin{figure}[ht!]
    \centering
    \includegraphics[width=\linewidth]{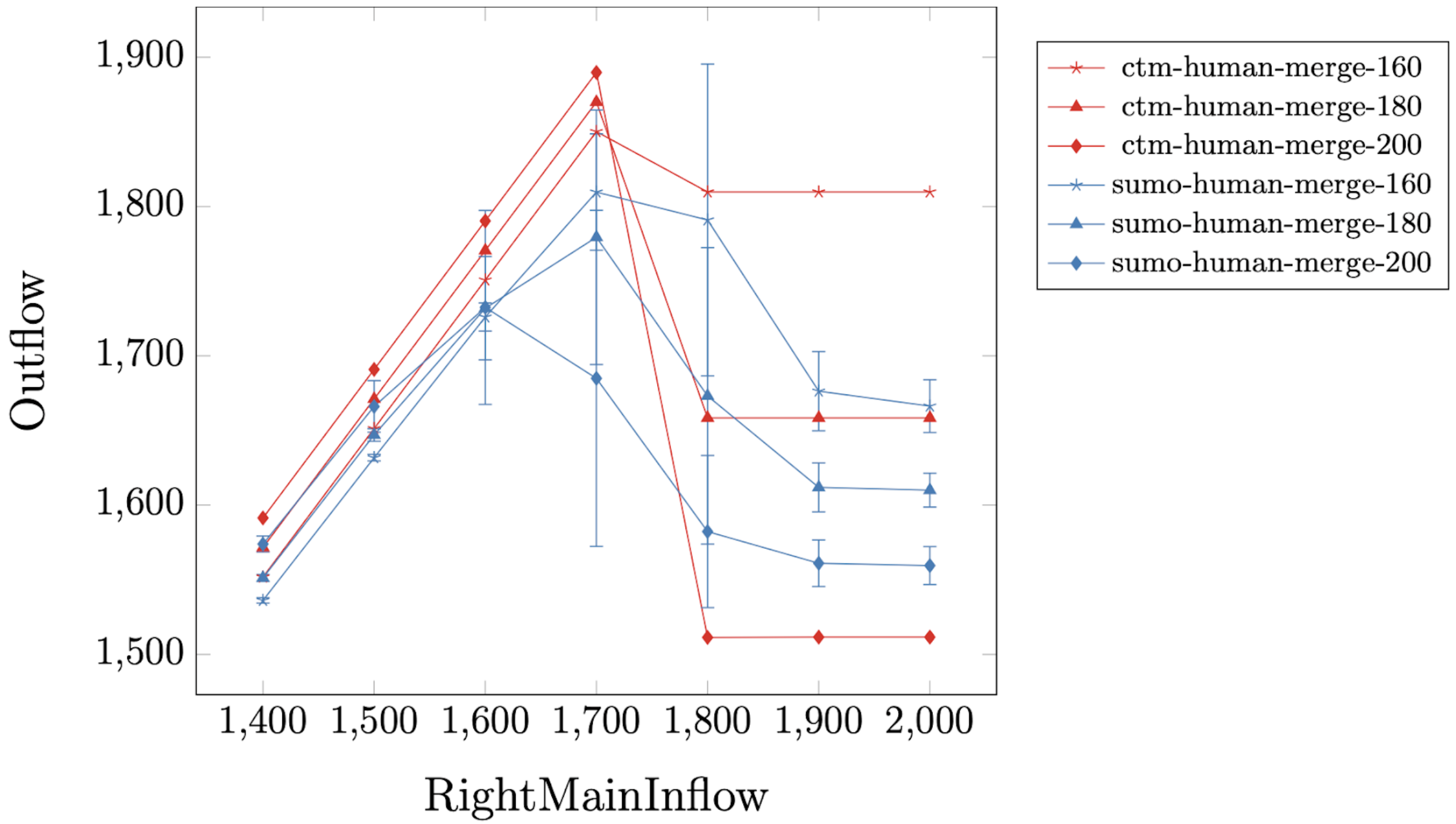}
    \caption{Comparing the inflow-outflow relation between SUMO and CTM under different main inflows and merge inflows. The range of the main inflow is [1400, 2000], and the range of the merge inflow is [160, 200]. 
    }
    \label{fig:simple_merge_ctm_outflow}
\end{figure}

To validate our novel single-lane merge CTM, we run a CTM simulation by iterating the operation suggested by Equation~(\ref{eq:ctm}) until the inflow and outflow converge to their steady state,
and then compare its overall inflow and outflow with SUMO's.
Figure~\ref{fig:simple_merge_ctm_outflow} shows this comparison, where each data point for SUMO is collected by running $100$ simulations, each  with a different random seed, and each data point for CTM is collected from a single simulation (since CTM is deterministic).  The CTM outflows  mostly fall within the 95\% confidence bounds of the mean, which represent 100 vehicles or fewer (around 5-6\% of the flow), thus providing reasonable similarity between  the inflow-outflow plots of the CTM and SUMO.
Both curves have similar values as the outflow first increase with inflow, then decreases as the traffic congestion develops, and finally saturates as we further increase the inflow. 

Running a CTM simulation takes less than a second, while running $100$ SUMO simulations can take minutes, or even hours or days for large scenarios. Therefore, CTM based on the triangular fundamental diagram can be viewed as a lower-fidelity but more  computationally efficient alternative for SUMO.

\subsection{Extending CTM to a double-lane merge scenario}
Next, we introduce another novel extension of CTM, modelling for the first time a multilane merge scenario. First, we discretize the double-lane scenario from Figure~\ref{fig:double_lane_road} into \SI{100}{\metre} cells, as illustrated in Figure~\ref{fig:ctm_double_lane}. Next, to capture traffic changing from neighboring cells, we add the following definitions:
\begin{figure}[ht!]
    \centering
    \includegraphics[width=\linewidth]{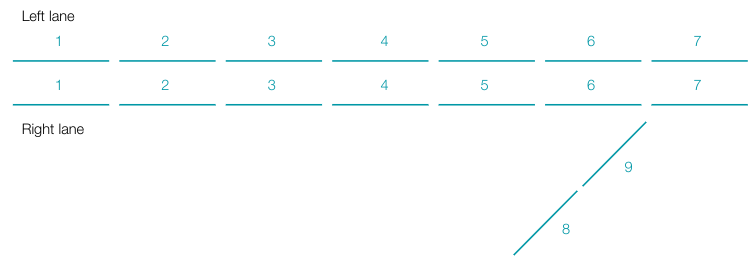}
    \caption{Discretizing the double-lane scenario from Figure~\ref{fig:double_lane_road} into CTM cells.}
    \label{fig:ctm_double_lane}
\end{figure}

\begin{itemize}
    \item $n_i^l(t), n_i^r(t)$: the number of vehicles on the left and right lanes of cell $i$ at time $t$ 
    \item ${lc}_i^l(t), {lc}_i^r(t)$: the number of lane-changes to the left and right lanes of cell $i$ at time $t$
\end{itemize}
We then add the following rules:
\begin{itemize}
    \item The right main road follows the same update rules as that of single-lane case.
    \item The left main road will not be blocked by the merging vehicles.
    \item Rules for lane-changing vehicles $lc^l_i(t)$ and $lc^r_i(t)$ from current lane to the target lane:
        \begin{itemize}
            \item If the number of vehicles in the current lane is less than or equal to that of the target lane, then more vehicles will be motivated to stay and the
            number of vehicles changing from current lane is small and denoted as $\epsilon$. 
            \item If the number of vehicles in the current lane is larger than that of the target lane, then additional vehicles will be motivated to change to the less congested lane. Here, we introduce a lane change factor $\delta$, to capture the fraction of vehicles that are motivated to change lanes: 
            \[
            lc_i^l(t) = \delta \times (n_i^r(t) - n_i^l(t)) + \epsilon
            \]
            \end{itemize}
    \item To capture the traffic congestion caused by lane-changing behaviors, we build flow discounting rules similar to those of the single-lane case as follows. If the number of vehicles changing to cell $i$ ($lc^l_i(t)$) is larger than 0 and the existing number of vehicles ($n^l_i(t)$) is larger than a certain threshold, then there will be congestion caused by lane changing and we discount the outflow using the previously introduced discounting factor $\alpha$: \[
            y_{i+1}(t) := \alpha \times y_{i+1}(t)
            \]
\end{itemize}

Based on the rules above, we can obtain a double-lane CTM, and a video of this model can be found: \url{https://youtu.be/cNzKiYEAkKQ}.

\begin{figure}[ht!]
\begin{subfigure}[bt]{\columnwidth}
    \centering
    \includegraphics[width=\linewidth]{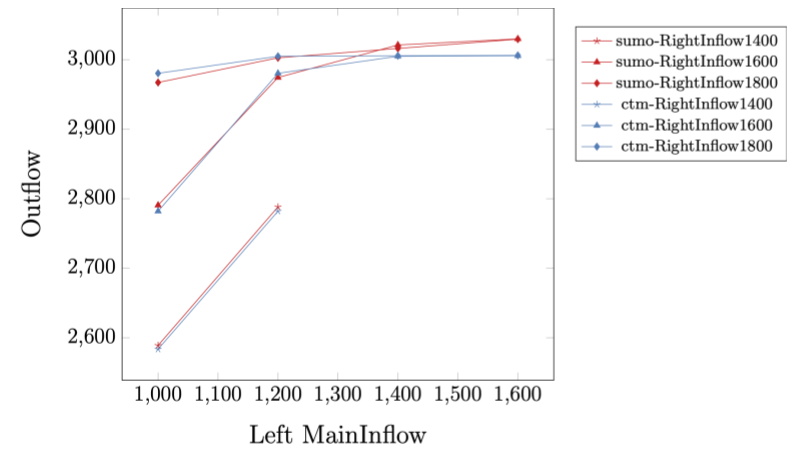}
    \caption{Fixing a few right-lane inflows, varying left-lane inflows\label{fig:ctm_left}}
\end{subfigure}
\begin{subfigure}[bt]{\columnwidth}
    \centering
    \includegraphics[width=\linewidth]{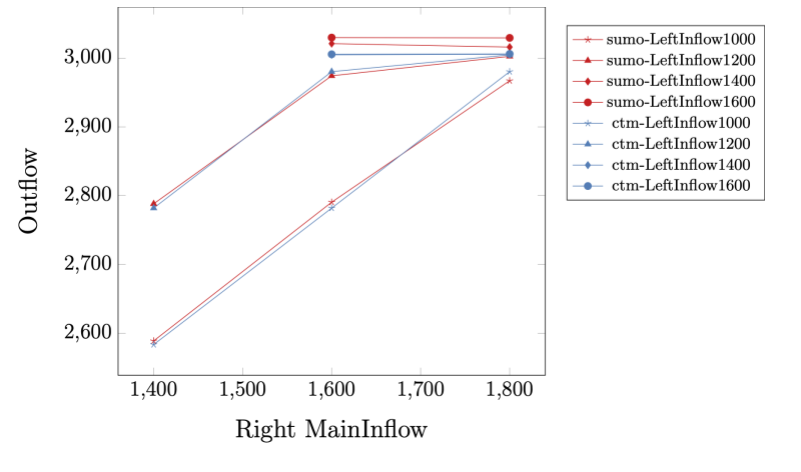}
    \caption{Fixing a few left-lane inflows, varying right-lane inflows\label{fig:ctm_right}}
\end{subfigure}
\caption{Comparing the inflows and outflows of the double-lane CTM with SUMO's. Here we only present the data points where the inflow on the left lane is smaller than that from the right lane. }
    \label{fig:ctm_double_lane_outflow}
\end{figure}

Similarly to the single-lane CTM, we validate the CTM by iterating the update equation until convergence of inflow and outflow, and then compare its overall inflow and outflow with SUMO's.
Figure~\ref{fig:ctm_double_lane_outflow} shows this comparison on a range of inflows and outflows, \changedyzz{where the main inflow on the right lane is chosen to be larger than that of the left lane so that most traffic changes from the right lane to the left lane to reflect a typical merge scenario.} 
It can be seen that the inflow-outflow curves match each other well. We conclude that the double-lane CTM that uses a triangular fundamental diagram to model each cell can serve as a lower-fidelity, computationally efficient alternative to SUMO for the double-lane merge scenario.

\commentdu{it would be good to put here links to the CTM videos that appear in slide 15}

\subsection{Insights from fundamental diagrams and CTM}
We introduced novel CTMs for single-lane and double-lane merge scenarios, by discretizing these roads into cells that are simulated using fitted triangular fundamental traffic flow diagrams. \changedyzz{We have observed that the inflow-outflow CTM plots approximate closely those of the SUMO micro-simulation, in both single-lane and double-lane merge scenarios. So the CTMs can be treated as a low-fidelity alternative of the SUMO microsimulator. In this section, we present insights about congestion reduction policies that are suggested by studying the behavior of our extended CTMs.}

\changedyz{In the triangular fundamental diagram shown in Figure~\ref{fig:triangular-fd}, the flow of each cell is maximized when the density is around a \emph{critical} density in which a maximal flow is achieved (the peak of the rectangle in Figure~\ref{fig:triangular-fd}). Hence,
it seems that an effective AV driving policy ought to seek to manipulate the traffic density in its vicinity to remain close to the critical density. Indeed, our proposed driving policy does so by slowing down to reduce traffic density if there is congestion ahead.}
\commentdu{can we claim that when the density in each cell is around the critical one, the overall flow is maximized? (we could check that by measuring cell flows when the overall outflow is maximized). If so, then (1) that might be one possible back-link, (2) we can say that future policies could be designed to explicitly manipulate densities towards the critical densities. }

\changedyz{
A similar intuition applies in the double-lane merge scenario as well. According to the lane-changing rules of CTM and SUMO, vehicles change from high-density lanes to low-density lanes. Autonomous vehicles are observed to encourage such lane-changing behaviors, by opening a gaps suitable for other cars to merge into. This behavior helps to optimize the traffic density in both lanes toward their critical densities.
}
The benefit of our extended CTMs could become even more apparent in large-scale multilane scenarios that are too slow to explore by exhaustive simulations of different traffic conditions. Using a similar approach, we can discretize such scenarios into cells modelled using fitted fundamental diagrams, and then use the computationally-efficiently CTMs to explore a range of traffic conditions and desired AV density control policies, which could direct the development of practical congestion reduction policies for large-scale scenarios.

\section{Implementation Details and Hyper-parameters}
\label{sec:implementation}
All experiments are built on top of SUMO 1.6.0 and UC Berkeley’s Flow software framework~\cite{wu2017flow}. The human-driven vehicles are controlled by the Krauss model with hyper-parameters defined in Table~\ref{hyperparameter2}. To control the autonomous vehicles, we use Proximal Policy Optimization algorithm~\cite{Schulman2017Proximal} to learn a driving policy, and the hyper-parameters for this algorithm is defined in Table~\ref{hyperparameter1}. The hyper-parameters used by CTM is shown in Table~\ref{hyperparameter3}. Our implementation is available at \url{https://github.com/yulinzhang/MITC-LARG}. 

\begin{table*}[h!]
  \centering
  \caption{Hyper-Parameters for Human-driven Vehicles}
  \label{hyperparameter2}
  \begin{tabular}{c|c}
     \textbf{Parameter} & \textbf{Value}\\
     \specialrule{.1em}{.05em}{.05em}
     \hline Controller & IDM Controller \\
     \hline Max Acceleration & 2.6\\
     \hline Max Deceleration & 4.5\\
     \hline Expected Time Headway & 1 second\\
     \hline
     \specialrule{.1em}{.05em}{.05em}
\end{tabular}
\end{table*}

\begin{table*}[h!]
  \centering
  \caption{Hyper-Parameters for Training Autonomous Vehicles}
  \label{hyperparameter1}
  \begin{tabular}{c|c}
     \textbf{Parameter} & \textbf{Value}\\
     \specialrule{.1em}{.05em}{.05em}
     Algorithm & Proximal Policy Optimization (PPO)\\
     \hline Horizon& 14000\\
     \hline Simulation Time Step Size & 0.5 \\
     \hline Optimizer & Stochastic Gradient Descent\\
     \hline Learning Rate & \shortstack{piece-wise linearly decreasing starting\\ from  $5\times10^{-4}$ (From scratch)} \\
     \hline Discount Factor ($\gamma$) & 0.998\\
     \hline GAE Lambda ($\lambda$) & 0.95\\
     \hline Actor Critic & True \\
     \hline Value Function Clip Parameter & $10^8$\\
     \hline Number of SGD Update per Iteration & 10 \\
     \hline Model hiddens & [100,50,25] \\
     \hline Clip Parameter & 0.2 \\
     \hline Entropy Coefficient & $10^{-3}$ \\
     \hline Sgd Minibatch size & 4096 \\
     \hline Train Batch Size & 60000\\
     \hline Value Function Share Layers & True \\
     \hline Value Loss Coefficient & 0.5 \\
     \hline KL Coefficient & 0.01 \\
     \hline KL Target & 0.01 \\
     \hline Max Acceleration & 2.6\\
     \hline Max Deceleration & 4.5\\
     \hline Training Iterations & 500\\
     \hline Number of Rollouts per Iteration & 30\\
     \hline Bonus & 20\\
     \hline $\eta$ & 0.9\\
     \hline
     \specialrule{.1em}{.05em}{.05em}
\end{tabular}
\end{table*}

\begin{table*}[h!]
  \centering
  \caption{Hyper-Parameters for the Extended Cell Transmission Model}
  \label{hyperparameter3}
  \begin{tabular}{c|c}
     \textbf{Parameter} & \textbf{Value}\\
     \specialrule{.1em}{.05em}{.05em}
     \hline $Q$ & \SI{4.0}{veh/s} \\
     \hline $N$ & \SI{14}{}\\
     \hline $v$ & \SI{21}{m/s}\\
     \hline $w$ & \SI{8.40}{m/s}\\
     \hline $d_c$ & \SI{0.04}{veh/m}\\
     \hline $\alpha$ & \SI{0.65}{}\\
     \hline $\beta$ & \SI{1}{}\\
     \hline $\delta$ & \SI{0.15}{}\\
     \hline $\epsilon$ & \SI{0.05}{}\\
     \hline
     \specialrule{.1em}{.05em}{.05em}
\end{tabular}
\end{table*}

\section{Conclusion and future work}
We  presented an approach for learning a congestion reduction driving policy that performs robustly in road merge scenarios over a variety of traffic conditions of practical interest. 
Specifically, the resulting policy reduces congestion in AV penetrations of \SI{1}{\percent}--\SI{40}{\percent}, traffic inflows ranging from no congestion to heavy congestion, random AV placement in traffic, 
single-lane single-merge road, single-lane road with two merges at varying distances, and double-lane single-merge road with lane changes.
The process of finding this policy involved identifying a single combination of
training conditions that yields a robust policy across different evaluating
conditions in a single-lane merge scenario. We find, for the first time, that
the resulting policy generalizes beyond the training conditions and road
geometry it was trained on.




Recently there has been an increasing interest in developing RL training methods
that result in robust policies. In our domain we find that randomizing AV
placement and searching for an effective training setup over the space of
traffic conditions achieve robustness effectively. The straightforward nature of
our method and its limited set of assumptions and tuning parameters make it a
potential candidate for real-world deployments. Given that RL algorithms have
been shown to be brittle in many domains, finding an RL-based policy that
performs robustly across a wide variety of traffic conditions in the challenging
domain of multiagent congestion reduction is both encouraging and somewhat
surprising. 

\changedyzz{As a secondary contribution of the article, and in order to more rapidly assess potential directions for reducing congestion at merge points, we introduced a novel variant of the Cell Transmission Model (CTM).  To this end, we first fit a fundamental diagram for the micro-simulation results in SUMO. Based on this fundamental diagram, we then construct an extended CTM that accounts for traffic congestion in the merge scenario. This extended CTM can serve as a lower fidelity, but more computationally efficient, alternative to micro-simulation, and can thus be leveraged for rapid prototyping. Additionally, we reflect on insights from experiments using the extended CTM model that motivate training policies that improve the traffic flow by keeping the traffic density close to the critical density from the fundamental diagram.}
\commentdu{do we have any evidence for the connection between critical density and max outflow, or is it a hypothesis? It is possible. But we have not yet investigated that. It is a hypothesis.} 

Nonetheless, our work has a few limitations that could serve as important directions for future research. \changedijcai{First, the question of whether there exists  a driving policy that reduces congestion when deployed on the left lane  of multilane scenarios still open.}
Second, our tests used the same aggressiveness level for all simulated human-driven vehicles. Testing with a variety of human behaviors would further increase the simulation results' applicability. Third, there is room to investigate a wider variety of road geometries beyond the ones we investigated.
Finally, even after investigating these extensions, there will likely be a sim2real gap to close, due to noisy/limited sensing and actuation delay. These limitations notwithstanding, this article’s contributions and insights advance our ongoing effort to reduce traffic congestion via AV control in the real world.

\section{Acknowledgement}
This work has taken place in the Learning Agents Research
Group (LARG) at the Artificial Intelligence Laboratory, The University
of Texas at Austin.  LARG research is supported in part by the
National Science Foundation (CPS-1739964, IIS-1724157, FAIN-2019844),
the Office of Naval Research (N00014-18-2243), Army Research Office
(W911NF-19-2-0333), DARPA, Lockheed Martin, General Motors, Bosch, and
Good Systems, a research grand challenge at the University of Texas at
Austin.  The views and conclusions contained in this document are
those of the authors alone.  Peter Stone serves as the Executive
Director of Sony AI America and receives financial compensation for
this work.  The terms of this arrangement have been reviewed and
approved by the University of Texas at Austin in accordance with its
policy on objectivity in research. 

\section{Compliance with Ethical Standards}
The prior and current affiliations that are in the conflict of interest include The University and Texas at Austin, General Motors, Texas A\&M University and Amazon Robotics. The corresponding author is prepared to collect documentation of compliance with ethical standards and send if requested.




\bibliography{flow_robust}


\end{document}